\documentclass[twocolumn, switch]{article} 

\usepackage{preprint}
\usepackage[dvipsnames, table]{xcolor}

\usepackage{amsmath, amsthm, amssymb, amsfonts}
\usepackage{bm}
\usepackage{graphicx}
\graphicspath{{img/}}
\usepackage{graphics}
\usepackage{enumerate}
\usepackage{setspace}
\usepackage{booktabs}
\usepackage{array}
\usepackage{tabularx}
\newcolumntype{Y}{>{\centering\arraybackslash}X}
\newcolumntype{C}[1]{>{\centering\arraybackslash}m{#1}}
\usepackage{wrapfig}
\usepackage{color}
\usepackage{multirow}
\usepackage{multicol}
\usepackage{threeparttable}
\usepackage{url}
\usepackage[page]{appendix}
\usepackage[numbers,square]{natbib}
\usepackage[utf8]{inputenc}
\usepackage[T1]{fontenc}
\usepackage{newunicodechar}
\newunicodechar{–}{--}
\newunicodechar{—}{---}
\newunicodechar{’}{'}
\usepackage{textcomp}
\usepackage{nicefrac}
\usepackage{microtype}
\usepackage{lineno}
\usepackage{float}
\usepackage{newfloat}
\DeclareFloatingEnvironment[name={Supplementary Figure}]{suppfigure}
\usepackage{sidecap}
\sidecaptionvpos{figure}{c}
\usepackage{subcaption}
\usepackage{algorithm}
\usepackage{algorithmic}
\usepackage[algoruled,algo2e]{algorithm2e}
\usepackage{titlesec}
\titlespacing\section{0pt}{12pt plus 3pt minus 3pt}{1pt plus 1pt minus 1pt}
\titlespacing\subsection{0pt}{10pt plus 3pt minus 3pt}{1pt plus 1pt minus 1pt}
\titlespacing\subsubsection{0pt}{8pt plus 3pt minus 3pt}{1pt plus 1pt minus 1pt}
\usepackage{hhline}
\usepackage{pifont}
\usepackage{colortbl}
\usepackage{stfloats}
\usepackage{balance}
\usepackage{etoolbox}
\usepackage[colorlinks = true,
            linkcolor = blue,
            urlcolor  = blue,
            citecolor = blue,
            anchorcolor = black]{hyperref}

\definecolor{Gray}{gray}{0.93}

\newcommand{\rev}[1]{#1}
\newcommand{\best}[1]{\textbf{#1}}
\newcommand{\second}[1]{\underline{#1}}
\newcommand{\thickhline}{\noalign{\global\arrayrulewidth=0.9pt}\hline\noalign{\global\arrayrulewidth=0.4pt}}
\newcommand{\thickcline}[1]{\noalign{\global\arrayrulewidth=0.9pt}\cline{#1}\noalign{\global\arrayrulewidth=0.4pt}}
\providecommand{\IEEEPARstart}[2]{#1#2}
\providecommand{\bstctlcite}[1]{}

\title{DiffCVE: Diffusion-based Compressed Video Enhancement}
\usepackage{authblk}

\author[1]{Wenqiang~Xiao}
\author[1]{Wenzhuo~Ma}
\author[1]{Junxi~Zhang}
\author[*1]{Zhenzhong~Chen}
\affil[1]{School of Remote Sensing and Information Engineering, Wuhan University}

\begin{document}

\twocolumn[ 
  \begin{@twocolumnfalse} 
  
\maketitle

\begin{abstract}
Perceptual quality enhancement of severely compressed videos remains challenging due to complex artifact patterns and substantial information loss. Recent diffusion models have demonstrated strong generative capability for visual restoration, but directly applying them to compressed video often ignores compression degradation characteristics and may introduce structure-inconsistent hallucinations. To address this issue, this paper presents a diffusion-based compressed video enhancement method, named DiffCVE. Coding Prior-enhanced Dual Conditioning (CPDC) branches are designed to jointly model compressed video and coding prior conditions, where coding priors including residuals and motion vectors provide complementary structural and motion guidance during the diffusion denoising process. To make the diffusion process aware of compression severity, a Compression Degradation Semantic Prompting (CDSP) mechanism is introduced to leverage QP-conditioned textual prompts together with LoRA fine-tuning. In addition, a Coding Prior-guided Weighted Fusion (CPWF) module is incorporated into the VAE decoder to fuse VAE encoder and coding prior encoder features with QP-predicted weights. Extensive experiments demonstrate the effectiveness of the proposed method in improving perceptual quality, especially under severe compression settings. \rev{The project page and code are available at \url{https://wqmaker.github.io/projects/DiffCVE/}.}
\end{abstract}

\vspace{0.4cm}

  \end{@twocolumnfalse} 
] 

\newcommand\blfootnote[1]{%
\begingroup
\renewcommand\thefootnote{}\footnote{#1}%
\addtocounter{footnote}{-1}%
\endgroup
}
\section{Introduction}
{\blfootnote{\scalebox{0.98}{Corresponding author: Zhenzhong Chen, E-mail:zzchen@ieee.org}}}
\IEEEPARstart{W}{ith} the rapid growth of online video platforms and streaming services, modern coding standards such as H.264/AVC~\cite{wiegand2003overview}, H.265/HEVC~\cite{sullivan2012overview}, and H.266/VVC~\cite{bross2021overview} are widely used to reduce storage and transmission costs. However, lossy compression, especially in low-bitrate scenarios, inevitably introduces blocking, ringing, blur, and structural distortion, which significantly impair perceptual quality and may harm downstream vision applications.

Existing compressed video enhancement methods have achieved notable progress. Early work mainly focused on single-frame or image enhancement~\cite{dong2015compression,wang2017novel}, while later approaches exploited temporal redundancy through motion compensation and temporal information aggregation from multiple frames~\cite{yang2018multi,peng2022ovqe,ding2023blind,jiang2023video}. More recently, transformer-based architectures have been explored to capture broader spatio-temporal dependencies~\cite{yu2023end,zhang2023video}.

Despite these advances, most existing methods are primarily optimized for distortion-oriented metrics such as PSNR and SSIM~\cite{wang2004image}. As a result, they often produce over-smoothed reconstructions that are not well aligned with human perception~\cite{yue2025survey}. To improve perceptual quality, recent works explore GAN-based approaches~\cite{wang2020multi,wang2021mw,chen2025perceptual} and frequency-guided strategies~\cite{dong2023temporal,peng2026hierarchical}. However, these methods still show limited capability in modeling the highly diverse distribution of compression artifacts, especially in severely compressed videos.

Diffusion models have recently emerged as a powerful class of generative models and have been successfully applied to image and video tasks~\cite{huang2024wavedm,yang2024motion,liu2025dm,wang2025seedvr}. Their strong generative prior makes them attractive for recovering realistic structures and textures from heavily degraded content. Nevertheless, directly applying diffusion models to compressed video quality enhancement still faces two key limitations. First, existing diffusion-based restoration methods usually treat compression distortion as a generic degradation and therefore have limited ability to adapt accurately to different compression levels and artifact patterns. Second, although diffusion models have strong texture generation capability, they may produce details inconsistent with the original content when effective structural constraints are absent, thereby impairing structural fidelity in restored results.

It is worth noting that rich coding-related cues are naturally available during video decoding, including residuals, motion vectors (MVs), and codec parameters such as the quantization parameter (QP). Specifically, residuals denote the prediction errors between the current frame and its predicted frame, which can reveal local reconstruction discrepancies and thus benefit spatial detail recovery. Motion vectors describe the displacement between matched blocks across frames, providing motion-strength information for preserving motion consistency. QP controls the quantization strength during compression and directly reflects compression severity, making it useful for characterizing rate-dependent degradation patterns. Although such information benefits various compressed-domain vision tasks~\cite{chen2021compressed,zhang2022codec,zhu2024deep,zhu2024cpga}, its joint integration with diffusion priors for compressed video perceptual enhancement remains under-explored.

Motivated by these observations, this paper proposes DiffCVE, a diffusion-based compressed video enhancement method. To the best of our knowledge, DiffCVE is the first compressed video enhancement framework to jointly exploit diffusion priors and coding priors. The main contributions are summarized as follows.

\begin{itemize}
    \item We propose Coding Prior-enhanced Dual Conditioning (CPDC) branches, a codec-aware conditioning design that jointly encodes compressed frames with coding priors. Residuals expose local prediction errors, while motion vectors provide motion strength cues. The resulting prior features both modulate compressed video condition generation and are injected as explicit conditions into the diffusion U-Net, enabling codec-derived spatial and motion information to guide restoration.
    \item We introduce a Compression Degradation Semantic Prompting (CDSP) mechanism that enables a single unified diffusion model to handle multiple QP levels. It converts QP-dependent compression characteristics into structured natural-language degradation descriptions and aligns them with the latent diffusion space through LoRA adaptation, providing semantic control across different compression severities.
    \item We develop a Coding Prior-guided Weighted Fusion (CPWF) module for the VAE decoder, where VAE decoder features, VAE encoder features, and coding-prior features are adaptively fused with QP-predicted weights. By adaptively fusing compressed video features, diffusion-generated features, and coding priors according to the compression severity, CPWF improves the balance between perceptual detail generation and content-faithful reconstruction.
\end{itemize}

The rest of the paper is organized as follows. Section~\ref{sec:related_work} reviews the relevant literature. Section~\ref{sec:method} presents the proposed method. Section~\ref{sec:experiments} provides the experimental results and analyses. Finally, Section~\ref{sec:conclusion} concludes the paper.

\section{Related Work}
\label{sec:related_work}

\subsection{Compressed Video Enhancement}
Existing compressed video enhancement methods can be broadly categorized into distortion-oriented and perceptual-oriented methods.

\textbf{Distortion-oriented methods:}
Distortion-oriented methods aim to optimize pixel-wise fidelity. Early methods focus on single-frame enhancement, such as ARCNN~\cite{dong2015compression}, which uses a convolutional network to suppress compression artifacts. Later methods incorporate temporal information. MFQE~\cite{yang2018multi} and MFQE 2.0~\cite{guan2019mfqe} leverage motion compensation to exploit higher-quality adjacent frames as references. STDR~\cite{luo2022spatio} improves spatio-temporal aggregation via multi-path deformable alignment.
More recently, transformer-based methods, including TVQE~\cite{yu2023end} and STCF~\cite{zhang2023video}, exploit attention-based modeling for spatio-temporal feature aggregation, while STFF~\cite{wang2025stff} further incorporates frequency-aware representations for enhanced reconstruction.

\textbf{Perceptual-oriented methods:}
Perceptual-oriented methods target human visual quality. GAN-based models are widely used to recover high-frequency details. MW-GAN~\cite{wang2020multi} introduces wavelet packet transform, while MW-GAN+~\cite{wang2021mw} further incorporates motion alignment and a 3D discriminator to enhance temporal coherence. HFHD-GAN~\cite{chen2025perceptual} leverages directional convolutions for high-frequency noise suppression. Beyond GANs, frequency-domain methods have also been explored. Dong \textit{et al.} propose a temporal wavelet transform-based method~\cite{dong2023temporal} to reduce computation via low-frequency enhancement, while HFGAT~\cite{peng2026hierarchical} enables cross-band interaction through independent and mixed frequency processing.

\subsection{Diffusion Models for Visual Enhancement}

Diffusion models have recently demonstrated strong performance in visual enhancement and restoration by leveraging powerful generative priors. Early works such as SR3~\cite{saharia2022image} and SRDiff~\cite{li2022srdiff} apply diffusion to image super-resolution, while later methods including StableSR~\cite{wang2024exploiting} and SeeSR~\cite{wu2024seesr} improve perceptual quality under real-world degradations.
Extending to video restoration, where temporal consistency is critical, MGLD-VSR~\cite{yang2024motion} introduces motion-guided diffusion sampling, DLoRAL~\cite{sun2026one} adopts dual-LoRA decoupling for one-step restoration. SeedVR~\cite{wang2025seedvr} and SeedVR2~\cite{wang2025seedvr2} use window-based diffusion transformers, with SeedVR2 enabling one-step restoration via adversarial post-training.
More recently, diffusion models have been explored for compression-oriented restoration. DriftRec~\cite{welker2024driftrec} modifies the forward SDE for compressed image restoration. SDATC~\cite{an2025spatial} combines spatial degradation-aware conditioning with temporal consistency modeling for compressed video super-resolution, while LViCAR~\cite{gehlot2025lvicar} uses diffusion priors for perceptual compression artifact reduction. These studies demonstrate the strong restoration capability of diffusion models across visual enhancement tasks, motivating us to explore diffusion-based compressed video enhancement for faithfully recovering details from heavily compressed videos.

\begin{figure*}[!t]
    \centering
    \includegraphics[width=1.0\textwidth]{model.pdf}
\caption{The architecture of the proposed DiffCVE framework. Coding priors (e.g., residuals and motion vectors) are incorporated to provide structure- and motion-aware guidance, while a large language model is used to generate QP-conditioned textual prompts for degradation semantics, which are further encoded as semantic embeddings to guide the diffusion process. During decoding, coding prior-guided weighted fusion is further applied to improve perceptual quality.}
\label{fig:framework}
\end{figure*}

\section{Method}
\label{sec:method}

The overall framework is illustrated in Fig.~\ref{fig:framework}. Given a low-quality compressed video sequence, this paper adapts a pretrained diffusion model to restore high-quality frames by injecting coding prior guidance and textual degradation semantics. Residual signals and motion vectors extracted from the codec provide structure- and motion-aware conditioning, while compression artifacts at different quantization levels are modeled as semantic embeddings for diffusion denoising. During decoding, temporal modeling and coding prior-guided feature fusion further improve frame consistency and perceptual quality.

\subsection{Coding Prior-enhanced Dual Conditioning Branches}
Compressed frames provide an important restoration basis, but using them alone may leave the diffusion model without enough structural and motion guidance. Coding priors, i.e., residual signals and motion vectors, provide complementary evidence from the compression process, thereby helping the model restore content more faithfully. Motivated by this, Coding Prior-enhanced Dual Conditioning branches are introduced, consisting of a compressed video conditioning branch and a coding prior conditioning branch that jointly generate coding prior conditioning features and coding prior-modulated video conditioning features for diffusion denoising.

\textbf{Latent Encoding:}
For a given compressed bitstream, the codec first decodes the compressed video $\mathbf{X}$. During this decoding process, the residuals $\mathbf{R}$ and motion vectors $\mathbf{V}$ are parsed from the bitstream.

To reduce pixel-space computation while preserving the main visual content, the compressed video is encoded into a compact latent representation using a pretrained VAE encoder:
\begin{equation}
z_{v} = E_{v}(\mathbf{X}),
\end{equation}

For the coding prior conditioning branch, residuals and motion vectors provide codec-side evidence about prediction errors and motion strength. To make such coding priors usable for latent condition generation, a lightweight coding prior encoder is designed to map them into a compact coding prior latent:
\begin{equation}
z_{p} = E_{p}(\mathbf{R}, \mathbf{V}),
\end{equation}
where $E_{v}(\cdot)$ and $E_{p}(\cdot)$ denote the pretrained VAE encoder and the lightweight coding prior encoder, respectively. The coding prior encoder is mainly composed of downsampling layers and residual blocks equipped with GroupNorm-SiLU-Conv (GSC) units. The resulting $z_{v}$ and $z_{p}$ serve as the compressed video latent and coding prior latent, respectively.
  
\textbf{Conditioning Generation:}
The compressed video latent preserves the main appearance and content layout, while its representation is inevitably entangled with compression artifacts. Therefore, a video condition generated solely from compressed frames may inherit distortion-biased and structurally unreliable responses. Residuals locate prediction errors, and motion vectors describe temporal correspondence, helping the video branch identify degradation-related regions and construct more structure-consistent visual conditions. However, even after such modulation, the generated video condition remains an implicit representation derived from degraded observations, and explicit codec-side cues may be weakened during condition construction. Therefore, coding priors are also retained as an independent condition for the diffusion U-Net, where they provide direct guidance throughout the iterative denoising process. To support these two roles, CPDC first uses a Coding Prior Conditioning Module (CPCM) to project the coding prior latent into two groups of conditional features:
\begin{equation}
\{f_{pv}^l\}, \{f_{pu}^l\} = G_{p}(z_{p}),
\end{equation}
where $G_{p}(\cdot)$ denotes CPCM and $l$ denotes the feature level. The group $\{f_{pv}^l\}$ modulates compressed video condition generation, while $\{f_{pu}^l\}$ is retained as an independent U-Net condition, allowing coding priors to both improve video condition reliability and provide direct codec-consistent denoising guidance.

The Compressed Video Conditioning Module (CVCM) $G_v(\cdot)$ takes $z_v$ as the main visual input and uses $\{f_{pv}^l\}$ to produce coding prior-modulated video conditioning features:
\begin{equation}
\{\tilde{f}_v^l\} = G_v(z_v,\{f_{pv}^l\}),
\end{equation}
where $\tilde{f}_v^l$ denotes the coding prior-modulated video conditioning feature at level $l$.

As shown in Fig.~\ref{fig:framework}, CVCM contains residual blocks and spatial attention layers for content representation, while coding priors are injected through Spatial Feature Transform (SFT)~\cite{wang2018recovering}. Instead of generating video conditions solely from the compressed video latent, SFT uses coding prior features to adaptively recalibrate the intermediate visual features. This allows codec-side evidence to guide condition generation and compensate for distortion-biased and structurally unreliable responses caused by compression artifacts. At each level $l$, SFT predicts spatially adaptive affine parameters from coding prior features:
\begin{equation}
\tilde{f}_v^l = (1 + \alpha^l)\odot f_v^l + \beta^l,\quad
\alpha^l,\beta^l = \mathcal{M}^l(f_{pv}^l),
\end{equation}
where $\alpha^l$ and $\beta^l$ are spatially adaptive scaling and shifting parameters, $\odot$ denotes element-wise multiplication, $\mathcal{M}^l(\cdot)$ maps coding prior features to affine parameters, and $f_v^l$ is the intermediate video conditioning feature before modulation. Thus, the generated video condition remains content-aware while being adjusted by residual and motion-vector evidence, leading to a more reliable condition for subsequent diffusion denoising.

Overall, CPDC outputs the coding prior-modulated video features $\{\tilde{f}_v^l\}$ and coding prior features $\{f_{pu}^l\}$. The former provides coding prior-corrected visual conditions, whereas the latter preserves explicit residual and motion cues for direct denoising guidance, helping the diffusion model recover perceptually plausible details with structural consistency and motion coherence.

\subsection{Compression Degradation Semantic Prompting Mechanism} 

Compressed videos encoded at different QP levels exhibit distinct degradation characteristics. Treating all compression levels uniformly limits enhancement quality, whereas training a separate model for each QP incurs high training and deployment costs. This paper therefore incorporates QP information into a unified model for multiple bitrate points.

Since Stable Diffusion is pretrained on large-scale text–image pairs, it has a strong capability to understand textual prompts, making it suitable for perceiving compression severity across different QP levels via text-guided conditioning. To exploit this capability, we use natural-language descriptions to characterize compression severity and artifact patterns, including blockiness, ringing, blur, and noise-like distortions. These semantic descriptions guide the pretrained diffusion model in restoring compressed videos.

\rev{Following previous compression settings~\cite{wang2021mw}, QP 32, 37, and 42 are used for training.} For each level, a textual prompt describes artifact characteristics and overall visual quality. \rev{As illustrated in Fig.~\ref{fig:framework}, a user queries a large language model (LLM), and its responses are used as fixed textual prompts for the corresponding QP levels.} These prompts are generated using the OpenAI GPT-5 model family~\cite{singh2025openai} to provide structured and perceptually meaningful descriptions of compression degradation, forming a semantic severity ordering where higher QP indicates stronger degradation, as listed in Table~\ref{tab:prompt}.

\rev{Considering the codec QP range and the suitability of diffusion models for heavily degraded inputs, we extend the inference range of CDSP to QP 51. Natural language prompts describe degradation at a semantic level and do not precisely distinguish every numerical QP. We therefore use anchor-based embedding interpolation, with additional prompts at QP 47 and 51 specified only for inference.}

\begin{table}[!t]
    \caption{Textual prompts for degradation semantics constructed for different QP levels.}
    \label{tab:prompt}
    \centering
    \footnotesize
    \renewcommand{\arraystretch}{1.1}
    \setlength{\tabcolsep}{3pt}
    \begin{tabular}{C{0.45cm} C{1.45cm} >{\raggedright\arraybackslash}m{0.65\columnwidth}}
    \toprule
    QP & Usage & \multicolumn{1}{c}{Textual prompt for degradation semantics} \\
    \midrule
    32 & Training \& inference & a moderately compressed video frame with mild blockiness and slight ringing, very slight blurring and minimal noise, medium visual quality \\
    37 & Training \& inference & a heavily compressed video frame with noticeable blockiness and ringing, slight blurring and minor noise-like artifacts, low visual quality \\
    42 & Training \& inference & a severely compressed video frame with severe blockiness and strong ringing, clear blurring and visible noise-like artifacts, very low visual quality \\
    \midrule
    47 & Inference only & an extremely compressed video frame with extreme blockiness and extremely strong ringing, severe blurring and strong noise-like artifacts, extremely poor visual quality \\
    51 & Inference only & an extremely compressed video frame with extreme blockiness and extremely strong ringing, extremely severe blurring and extremely strong noise-like artifacts, extremely poor visual quality \\
    \bottomrule
    \end{tabular}
\end{table} 

\rev{As shown in Fig.~\ref{fig:framework}, for each anchor QP $q \in \{32,37,42,47,51\}$, the corresponding textual prompt is encoded using a pretrained CLIP text encoder~\cite{radford2021learning}:}
\begin{equation}
z_{qp}^{q} = E_{\text{text}}(T_{qp}^{q}),
\end{equation}
where $T_{qp}^{q}$ denotes the textual prompt for degradation semantics corresponding to the QP level $q$, $z_{qp}^{q}$ denotes the corresponding embedding for degradation semantics, and $E_{\text{text}}$ denotes the pretrained CLIP text encoder.

\rev{All anchor prompts are encoded offline. For an input QP $q$ lying between two adjacent anchor QPs $q_l$ and $q_h$, we linearly interpolate their embeddings:
\begin{equation}
\begin{aligned}
z_{qp} &= (1-\alpha)z_{qp}^{q_l}+\alpha z_{qp}^{q_h}, \\
\alpha &= \frac{q-q_l}{q_h-q_l}.
\end{aligned}
\label{eq:qp_anchor_interpolation}
\end{equation}
Here, $q_l\leq q\leq q_h$. When $q$ coincides with an anchor QP, its embedding is used directly. This provides a continuous QP-dependent semantic condition without retraining the model for the additional inference QPs.}

Overall, CDSP produces the degradation semantic embedding $z_{qp}$ for the subsequent diffusion denoising process. This condition provides QP-dependent semantic guidance about compression severity and artifact characteristics, enabling a unified diffusion model to adjust its restoration behavior across different bitrate levels.

\subsection{Conditioning Injection into the Diffusion U-Net}

After CPDC and CDSP generate their conditions, the coding prior-modulated video conditioning features $\{\tilde{f}_v^l\}$, coding prior conditioning features $\{f_{pu}^l\}$, and degradation semantic embedding $z_{qp}$ are jointly injected into the diffusion U-Net, as illustrated in Fig.~\ref{fig:framework}. The denoising process is formulated as
\begin{equation}
\hat{\epsilon} = \epsilon_\theta(z_t, t, \{\tilde{f}_v^l\}, \{f_{pu}^l\}, z_{qp}),
\label{eq:conditioned_denoise} 
\end{equation}
where $z_t$ is the noisy latent at timestep $t$ in the reverse diffusion trajectory from $z_T$ to $z_0$, $\epsilon_\theta$ denotes the diffusion denoising network, and $\hat{\epsilon}$ is the predicted noise.

The video and coding prior conditioning features are injected into multiple U-Net levels through SFT-based modulation to exploit structural discrepancy and motion-aware coding cues, while $z_{qp}$ is injected through cross-attention layers to provide compression severity and artifact semantics.

Although pretrained diffusion models can understand textual prompts, they cannot effectively associate text embeddings with degradation representations of different severities in the latent space. Therefore, targeted fine-tuning is required to better align degradation semantics with latent representations. To this end, LoRA~\cite{hu2022lora} is introduced to efficiently adapt the diffusion model. This enables the diffusion model to align its latent representation with degradation semantics while preserving pretrained generative priors.

\subsection{Coding Prior-guided Weighted Fusion Module}
Stable Diffusion is originally trained on images, so extending it to videos may cause temporal discontinuity. Following~\cite{yang2024motion}, temporal convolution and temporal attention blocks are inserted into the middle block of the diffusion U-Net to model temporal dynamics during the diffusion sampling process, as shown in Fig.~\ref{fig:framework}. However, since diffusion is performed in the low-resolution latent space, video reconstruction through the VAE decoder may still introduce pixel-level inter-frame flickering artifacts. To mitigate this issue, a temporal convolution is incorporated into the decoder to capture local temporal dynamics across frames~\cite{yang2024motion}, as illustrated in Fig.~\ref{fig:framework}.

For the intermediate decoder feature $f_d$ at different levels of the VAE decoder, temporal convolution is first applied to enhance local temporal dynamics:
\begin{equation}
\hat{f}_d = \mathcal{T}(f_d),
\end{equation}
where $\mathcal{T}(\cdot)$ denotes temporal convolution.

While diffusion models possess strong generative capability, their outputs may deviate from the ground-truth signal, causing reduced fidelity and structure-inconsistent details. To address this issue, previous work~\cite{wang2024exploiting} proposes Controllable Feature Wrapping, which fuses diffusion features with VAE encoder features to anchor decoder reconstruction to the input layout.

\begin{table*}[!t]
    \caption{Full-reference quantitative comparison. The metrics are reported in the order of $\Delta$PSNR$\uparrow$ / $\Delta$SSIM$\uparrow$ / $\Delta$LPIPS$\downarrow$ / $\Delta$DISTS$\downarrow$. Rows correspond to QP levels, classes, and sequence indices, while columns are grouped by method and metric. QP 42 sequence results are followed by QP 42/37/32 average results. Best and second-best values are marked in bold and underlined, respectively.}
    \label{tab:compare_fr}
    \centering
    \footnotesize
    \renewcommand{\arraystretch}{1.8}
    \setlength{\tabcolsep}{0pt}
    \setlength{\arrayrulewidth}{0.4pt}
    \resizebox{\textwidth}{!}{%
    \begin{tabular}{!{\vrule width 0.9pt}C{0.46cm}!{\vrule width 0.9pt}C{0.70cm}!{\vrule width 0.9pt}C{1.12cm}!{\vrule width 0.9pt}C{1.15cm}|C{1.15cm}|C{1.15cm}|C{1.15cm}!{\vrule width 0.9pt}C{1.15cm}|C{1.15cm}|C{1.15cm}|C{1.15cm}!{\vrule width 0.9pt}C{1.15cm}|C{1.15cm}|C{1.15cm}|C{1.15cm}!{\vrule width 0.9pt}C{1.15cm}|C{1.15cm}|C{1.15cm}|C{1.15cm}!{\vrule width 0.9pt}C{1.15cm}|C{1.15cm}|C{1.15cm}|C{1.15cm}!{\vrule width 0.9pt}C{1.15cm}|C{1.15cm}|C{1.15cm}|C{1.15cm}!{\vrule width 0.9pt}C{1.15cm}|C{1.15cm}|C{1.15cm}|C{1.15cm}!{\vrule width 0.9pt}}
    \thickhline
    \multirow[c]{2}{*}{QP} & \multirow[c]{2}{*}{Class} & \multirow[c]{2}{*}{\scalebox{0.92}{Sequence}} & \multicolumn{4}{c!{\vrule width 0.9pt}}{{\footnotesize\renewcommand{\arraystretch}{1.0}\begin{tabular}{@{}c@{}}\rule{0pt}{2.4ex}STDR\\[0.25ex](TMM 2023)\end{tabular}}} & \multicolumn{4}{c!{\vrule width 0.9pt}}{{\footnotesize\renewcommand{\arraystretch}{1.0}\begin{tabular}{@{}c@{}}\rule{0pt}{2.4ex}STFF\\[0.25ex](TBC 2025)\end{tabular}}} & \multicolumn{4}{c!{\vrule width 0.9pt}}{{\footnotesize\renewcommand{\arraystretch}{1.0}\begin{tabular}{@{}c@{}}\rule{0pt}{2.4ex}MW-GAN+\\[0.25ex](TCSVT 2021)\end{tabular}}} & \multicolumn{4}{c!{\vrule width 0.9pt}}{{\footnotesize\renewcommand{\arraystretch}{1.0}\begin{tabular}{@{}c@{}}\rule{0pt}{2.4ex}HFGAT\\[0.25ex](AAAI 2026)\end{tabular}}} & \multicolumn{4}{c!{\vrule width 0.9pt}}{{\footnotesize\renewcommand{\arraystretch}{1.0}\begin{tabular}{@{}c@{}}\rule{0pt}{2.4ex}SeedVR2-3B\\[0.25ex](ICLR 2026)\end{tabular}}} & \multicolumn{4}{c!{\vrule width 0.9pt}}{{\footnotesize\renewcommand{\arraystretch}{1.0}\begin{tabular}{@{}c@{}}\rule{0pt}{2.4ex}DiffCVE-50\\[0.25ex](Ours)\end{tabular}}} & \multicolumn{4}{c!{\vrule width 0.9pt}}{{\footnotesize\renewcommand{\arraystretch}{1.0}\begin{tabular}{@{}c@{}}\rule{0pt}{2.4ex}DiffCVE-1\\[0.25ex](Ours)\end{tabular}}} \\
    \thickcline{4-31}
    & & & {\footnotesize PSNR} & {\footnotesize SSIM} & {\footnotesize LPIPS} & {\footnotesize DISTS} & {\footnotesize PSNR} & {\footnotesize SSIM} & {\footnotesize LPIPS} & {\footnotesize DISTS} & {\footnotesize PSNR} & {\footnotesize SSIM} & {\footnotesize LPIPS} & {\footnotesize DISTS} & {\footnotesize PSNR} & {\footnotesize SSIM} & {\footnotesize LPIPS} & {\footnotesize DISTS} & {\footnotesize PSNR} & {\footnotesize SSIM} & {\footnotesize LPIPS} & {\footnotesize DISTS} & {\footnotesize PSNR} & {\footnotesize SSIM} & {\footnotesize LPIPS} & {\footnotesize DISTS} & {\footnotesize PSNR} & {\footnotesize SSIM} & {\footnotesize LPIPS} & {\footnotesize DISTS} \\
    \thickhline
    \footnotesize
    \multirow[c]{19}{*}{42} & \multirow[c]{2}{*}{A} & 1 & 0.770 & \second{0.027} & -0.003 & 0.022 & \best{0.947} & \best{0.028} & 0.000 & 0.033 & -0.354 & -0.020 & -0.050 & -0.034 & \second{0.806} & 0.009 & -0.075 & -0.043 & -3.969 & -0.073 & 0.062 & 0.024 & -0.206 & -0.018 & \second{-0.085} & \second{-0.052} & -0.451 & -0.051 & \best{-0.094} & \best{-0.070} \\
    \cline{3-31}
     &  & 2 & 0.305 & \second{0.022} & 0.001 & 0.017 & \best{0.435} & \best{0.024} & 0.008 & 0.022 & -0.364 & -0.033 & -0.128 & -0.069 & \second{0.429} & 0.002 & -0.138 & -0.058 & -2.817 & -0.048 & -0.035 & -0.046 & -0.240 & -0.038 & \best{-0.147} & \second{-0.092} & -0.364 & -0.052 & \second{-0.143} & \best{-0.109} \\
    \thickcline{2-31}
     & \multirow[c]{5}{*}{B} & 3 & 0.392 & \second{0.025} & -0.003 & 0.029 & \best{0.672} & \best{0.026} & 0.005 & 0.037 & -0.095 & -0.015 & -0.119 & \best{-0.069} & \second{0.422} & -0.001 & -0.110 & -0.036 & -1.702 & -0.020 & -0.052 & -0.011 & -0.127 & -0.035 & \second{-0.136} & -0.048 & -0.158 & -0.038 & \best{-0.142} & \second{-0.051} \\
    \cline{3-31}
     &  & 4 & 0.325 & \second{0.019} & -0.005 & 0.009 & \second{0.473} & \best{0.021} & -0.003 & 0.014 & -0.080 & -0.032 & -0.131 & \best{-0.074} & \best{0.520} & -0.013 & -0.134 & -0.055 & -1.544 & -0.004 & -0.101 & -0.054 & 0.068 & -0.021 & \best{-0.149} & -0.064 & -0.338 & -0.058 & \second{-0.148} & \second{-0.065} \\
    \cline{3-31}
     &  & 5 & 0.374 & \best{0.022} & 0.001 & 0.001 & \best{0.493} & \second{0.021} & 0.004 & 0.006 & -0.392 & -0.023 & \second{-0.123} & -0.053 & \second{0.383} & 0.002 & -0.110 & -0.044 & -2.231 & -0.008 & -0.081 & -0.050 & -0.262 & -0.003 & \best{-0.137} & \best{-0.067} & -0.611 & -0.027 & \best{-0.137} & \second{-0.065} \\
    \cline{3-31}
     &  & 6 & 0.292 & \second{0.021} & -0.004 & 0.010 & \best{0.650} & \best{0.026} & -0.006 & 0.012 & -0.159 & -0.032 & -0.136 & \best{-0.065} & \second{0.428} & -0.009 & -0.117 & -0.036 & -1.705 & -0.014 & -0.056 & -0.058 & 0.164 & -0.009 & \best{-0.155} & -0.060 & 0.093 & -0.015 & \second{-0.140} & \second{-0.063} \\
    \cline{3-31}
     &  & 7 & \second{0.239} & \second{0.024} & 0.021 & 0.029 & \best{0.388} & \best{0.026} & 0.027 & 0.034 & -0.469 & -0.037 & -0.168 & \best{-0.075} & 0.178 & -0.007 & \second{-0.169} & -0.059 & -1.703 & -0.004 & -0.090 & -0.062 & -0.344 & -0.032 & \best{-0.193} & -0.060 & -0.452 & -0.050 & -0.155 & \second{-0.074} \\
    \thickcline{2-31}
     & \multirow[c]{4}{*}{C} & 8 & \second{0.416} & \best{0.028} & 0.007 & 0.003 & 0.296 & \best{0.028} & 0.013 & 0.014 & -0.333 & -0.031 & -0.077 & -0.063 & \best{0.456} & \second{0.005} & \best{-0.101} & -0.058 & -2.354 & -0.024 & -0.038 & -0.057 & -0.073 & -0.011 & -0.097 & \second{-0.064} & -0.105 & -0.014 & \second{-0.099} & \best{-0.073} \\
    \cline{3-31}
     &  & 9 & 0.504 & \second{0.031} & -0.020 & -0.003 & \second{0.602} & \best{0.032} & -0.019 & 0.002 & -0.020 & -0.022 & -0.120 & -0.072 & \best{0.852} & 0.005 & \second{-0.170} & -0.072 & -1.822 & -0.004 & -0.112 & -0.072 & -0.404 & -0.092 & -0.155 & \second{-0.077} & -0.624 & -0.119 & \best{-0.182} & \best{-0.108} \\
    \cline{3-31}
     &  & 10 & \second{0.237} & \best{0.032} & -0.003 & -0.004 & 0.211 & \best{0.032} & 0.007 & 0.010 & -0.705 & -0.040 & -0.056 & -0.054 & \best{0.256} & \second{0.010} & -0.119 & -0.055 & -2.388 & -0.029 & -0.003 & -0.029 & -0.413 & -0.022 & \second{-0.120} & \second{-0.071} & -0.476 & -0.018 & \best{-0.125} & \best{-0.085} \\
    \cline{3-31}
     &  & 11 & \second{0.194} & \best{0.020} & 0.030 & 0.013 & \best{0.293} & \best{0.020} & 0.036 & 0.024 & -0.427 & -0.043 & -0.095 & -0.063 & 0.146 & \second{-0.009} & -0.119 & -0.060 & -2.303 & -0.016 & -0.032 & -0.040 & -0.697 & -0.042 & \best{-0.130} & \second{-0.085} & -0.726 & -0.045 & \second{-0.128} & \best{-0.091} \\
    \thickcline{2-31}
     & \multirow[c]{4}{*}{D} & 12 & \best{0.519} & \best{0.025} & -0.006 & -0.004 & 0.152 & \second{0.022} & 0.007 & 0.007 & -0.480 & -0.023 & -0.062 & -0.067 & \second{0.321} & 0.003 & \second{-0.109} & -0.055 & -3.013 & -0.054 & 0.016 & -0.016 & -0.270 & -0.015 & -0.103 & \second{-0.087} & -0.691 & -0.059 & \best{-0.126} & \best{-0.095} \\
    \cline{3-31}
     &  & 13 & 0.412 & \second{0.025} & 0.011 & 0.006 & \best{0.493} & \best{0.026} & 0.015 & 0.014 & -0.408 & -0.037 & -0.096 & -0.068 & \second{0.444} & -0.003 & \second{-0.117} & -0.054 & -2.370 & -0.015 & -0.068 & -0.066 & -0.567 & -0.055 & \second{-0.117} & \best{-0.099} & -0.655 & -0.071 & \best{-0.129} & \second{-0.090} \\
    \cline{3-31}
     &  & 14 & \second{0.224} & \best{0.032} & 0.014 & -0.011 & 0.169 & \second{0.031} & 0.033 & 0.004 & -0.583 & -0.026 & -0.089 & -0.072 & \best{0.225} & 0.014 & \second{-0.137} & -0.059 & -1.811 & -0.011 & -0.017 & -0.049 & -0.500 & -0.015 & \best{-0.158} & \best{-0.111} & -0.593 & -0.016 & \best{-0.158} & \second{-0.110} \\
    \cline{3-31}
     &  & 15 & \second{0.325} & \best{0.025} & 0.024 & 0.009 & \best{0.416} & \best{0.025} & 0.028 & 0.019 & -0.424 & -0.039 & -0.088 & -0.063 & 0.216 & \second{-0.002} & -0.114 & -0.061 & -2.765 & -0.018 & -0.044 & -0.053 & -0.658 & -0.038 & \second{-0.122} & \second{-0.107} & -0.702 & -0.044 & \best{-0.137} & \best{-0.113} \\
    \thickcline{2-31}
     & \multirow[c]{3}{*}{E} & 16 & 0.665 & \second{0.023} & -0.032 & -0.020 & \second{0.824} & \best{0.025} & -0.030 & -0.010 & -0.265 & -0.019 & -0.095 & \second{-0.084} & \best{0.838} & 0.009 & \best{-0.110} & -0.075 & -4.435 & -0.043 & -0.029 & -0.073 & -0.432 & -0.013 & \second{-0.098} & \best{-0.087} & -0.754 & -0.025 & -0.089 & -0.083 \\
    \cline{3-31}
     &  & 17 & 0.437 & \best{0.010} & -0.013 & 0.002 & \second{0.510} & \best{0.010} & -0.012 & 0.014 & 0.258 & -0.011 & -0.090 & \best{-0.086} & \best{0.701} & \second{0.002} & -0.094 & -0.057 & -1.938 & -0.010 & -0.036 & \second{-0.078} & -0.096 & -0.007 & \second{-0.105} & -0.075 & -0.531 & -0.022 & \best{-0.113} & \best{-0.086} \\
    \cline{3-31}
     &  & 18 & 0.748 & \second{0.018} & -0.019 & -0.001 & \second{0.859} & \best{0.019} & -0.016 & 0.012 & -0.033 & -0.018 & -0.101 & -0.079 & \best{0.888} & 0.002 & -0.101 & -0.065 & -2.732 & -0.014 & -0.072 & \second{-0.086} & -0.841 & -0.022 & \best{-0.116} & -0.077 & -1.066 & -0.036 & \second{-0.110} & \best{-0.087} \\
    \thickcline{2-31}
     & \multicolumn{2}{c!{\vrule width 0.9pt}}{Avg.} & 0.410 & \second{0.024} & 0.000 & 0.006 & \best{0.493} & \best{0.025} & 0.005 & 0.015 & -0.296 & -0.028 & -0.101 & -0.067 & \second{0.473} & 0.001 & -0.119 & -0.056 & -2.422 & -0.023 & -0.044 & -0.049 & -0.328 & -0.027 & \second{-0.129} & \second{-0.077} & -0.511 & -0.042 & \best{-0.131} & \best{-0.084} \\
    \thickhline
    \multirow[c]{1}{*}{37} & \multicolumn{2}{c!{\vrule width 0.9pt}}{Avg.} & \second{0.300} & \best{0.018} & 0.008 & 0.009 & \best{0.554} & \best{0.018} & 0.010 & 0.014 & -0.382 & -0.013 & -0.070 & -0.051 & 0.185 & \second{-0.002} & \best{-0.089} & -0.043 & -5.065 & -0.072 & 0.021 & -0.010 & -1.007 & -0.028 & \best{-0.089} & \second{-0.056} & -1.057 & -0.031 & \second{-0.085} & \best{-0.058} \\
    \thickhline
    \multirow[c]{1}{*}{32} & \multicolumn{2}{c!{\vrule width 0.9pt}}{Avg.} & \second{0.171} & \best{0.013} & 0.012 & 0.011 & \best{0.376} & \second{0.011} & 0.014 & 0.015 & -0.440 & -0.012 & \second{-0.049} & \best{-0.039} & -0.144 & -0.003 & \best{-0.062} & -0.032 & -7.877 & -0.124 & 0.081 & 0.027 & -2.197 & -0.035 & -0.047 & \second{-0.036} & -2.128 & -0.033 & -0.040 & -0.035 \\
    \thickhline
    \end{tabular}}
    \par\vspace*{3pt}
    \noindent\begin{minipage}{\textwidth}
    \raggedright
    \scriptsize
    The resolution of class A is $2560\times1600$, 1: PeopleOnStreet; 2: Traffic.\\
    The resolution of class B is $1920\times1080$, 3: BQTerrace; 4: BasketballDrive; 5: Cactus; 6: Kimono1; 7: ParkScene.\\
    The resolution of class C is $832\times480$, 8: BQMall; 9: BasketballDrill; 10: PartyScene; 11: RaceHorses.\\
    The resolution of class D is $416\times240$, 12: BQSquare; 13: BasketballPass; 14: BlowingBubbles; 15: RaceHorses.\\
    The resolution of class E is $1280\times720$, 16: FourPeople; 17: Johnny; 18: KristenAndSara.\\
    \end{minipage}
\end{table*}

For compressed video enhancement, we adapt this feature-wrapping idea by additionally incorporating coding prior features. Although VAE encoder features provide useful content-layout information, they are extracted from compressed frames and remain entangled with compression artifacts. Residuals and motion vectors provide complementary prediction-error and temporal-correspondence cues, helping the decoder handle remaining structure-inconsistent details after diffusion denoising. Specifically, the temporally enhanced decoder feature is concatenated with the VAE encoder feature $f_e$ and coding prior encoder feature $f_p$, and then passed through a small convolutional network $\mathcal{F}(\cdot)$:
\begin{equation}
f_{\text{conv}} = \mathcal{F}([\hat{f}_d, f_e, f_p]),
\end{equation}
where $\mathcal{F}(\cdot)$ consists of convolutional layers and Residual Dense Blocks (RDBs)~\cite{zhang2018residual}, and $[\cdot]$ denotes channel-wise concatenation. The final enhanced decoder feature is obtained via a weighted combination:
\begin{equation}
f_{\text{out}} = w \cdot f_{\text{conv}} + \hat{f}_d,
\end{equation}
where $w \in [0,1]$ is the adaptive fusion weight.

Instead of being manually set as in previous work, $w$ is predicted from QP using a lightweight MLP:
\begin{equation}
w = \sigma(\mathrm{MLP}(QP)),
\end{equation}
where $\sigma(\cdot)$ is the sigmoid function.

In this way, the CPWF module first enhances decoder features with temporal information and then integrates them with VAE encoder and coding prior encoder features at multiple decoder levels, while dynamically modulating feature fusion according to compression severity. The decoder therefore receives both content-layout and codec-side constraints for improving local fidelity and structure consistency.

\section{Experiments}
\label{sec:experiments}

\subsection{Experimental Setup}

\textbf{Datasets:} The proposed model is trained on Vimeo-90k~\cite{xue2019video}, where each sample contains seven consecutive frames with a resolution of $448\times256$. Paired data are generated via H.264 compression under Low Delay P at multiple bitrates. Evaluation uses 18 standard JCT-VC~\cite{ohm2012comparison} sequences. \rev{Additionally, we evaluate our model on the LDV Track 1 test dataset~\cite{yang2022ntire} and on H.265 compressed JCT-VC Class C and E sequences.}

\textbf{Metrics:} Four full-reference metrics, PSNR, SSIM~\cite{wang2004image}, LPIPS~\cite{zhang2018unreasonable}, and DISTS~\cite{ding2020image} are used. PSNR and SSIM measure reconstruction fidelity, while LPIPS and DISTS reflect perceptual quality. Higher values are better for PSNR and SSIM, while lower values are better for LPIPS and DISTS. We further evaluate perceptual quality using four no-reference metrics, MUSIQ~\cite{ke2021musiq}, CLIPIQA~\cite{wang2023exploring}, DOVER~\cite{wu2023exploring}, and NIQE~\cite{mittal2012making}, where higher values are better for MUSIQ, CLIPIQA, and DOVER, while lower values are better for NIQE. Temporal consistency is measured by tOF~\cite{chu2020learning}, where lower values indicate better consistency. In the subsequent analysis, for a metric $M$, $\Delta M$ denotes the difference between the enhanced result and the compressed input:
\begin{equation}
\Delta M = M_{\text{enhanced}} - M_{\text{compressed}},
\end{equation}
where $M_{\text{enhanced}}$ and $M_{\text{compressed}}$ denote the corresponding metric values. \rev{In the following tables and figures, we use $\uparrow$ and $\downarrow$ to indicate that higher and lower values are better, respectively.} All metrics are computed in the RGB color space. Since the target task is compressed video perceptual enhancement, this paper places particular emphasis on perceptual quality and temporal stability in the subsequent analysis.

\textbf{Implementation details:} \rev{We propose two variants, DiffCVE-50 and DiffCVE-1, both initialized from the same pretrained Stable Diffusion v2.1 backbone and weights. The two variants share most training configurations but follow different training procedures.} \rev{DiffCVE-50 is trained} on four NVIDIA GeForce RTX 4060 Ti GPUs in a two-stage pipeline. In the first stage, the diffusion U-Net and conditioning branches are fine-tuned using clips of $5\times256\times256$ with a batch size of 16. In the second stage, latent features are extracted from the trained diffusion model and used to train decoder-side modules with a batch size of 4.

For the single-step variant DiffCVE-1, we train the model at a fixed diffusion timestep using paired original and compressed videos. Due to limited computational resources, we adopt a progressive training strategy. The U-Net is first trained on four NVIDIA GeForce RTX 3090 GPUs, followed by decoder training on the same setup. \rev{The single-step formulation of DiffCVE-1 enables us to introduce an additional end-to-end joint training stage beyond the two-stage procedure used for DiffCVE-50. }Finally, the U-Net and decoder are jointly fine-tuned on two NVIDIA Tesla A100 40GB GPUs.

During inference, DiffCVE-50 uses 50 diffusion steps while DiffCVE-1 uses a single step. For both variants, full frames are directly processed for videos below 720p. For videos at 720p and above, overlapping patches are enhanced and then merged by weighted averaging.

\begin{table*}[!t]
    \caption{No-reference quantitative comparison. The metrics are reported in the order of $\Delta$MUSIQ$\uparrow$ / $\Delta$CLIPIQA$\uparrow$ / $\Delta$DOVER$\uparrow$ / $\Delta$NIQE$\downarrow$. Rows correspond to QP levels, classes, and sequence indices, while columns are grouped by method and metric. QP 42 sequence results are followed by QP 42/37/32 average results. Best and second-best values are marked in bold and underlined, respectively.}
    \label{tab:compare_nr}
    \centering
    \footnotesize
    \renewcommand{\arraystretch}{1.8}
    \setlength{\tabcolsep}{0pt}
    \setlength{\arrayrulewidth}{0.4pt}
    \resizebox{\textwidth}{!}{%
    \begin{tabular}{!{\vrule width 0.9pt}C{0.46cm}!{\vrule width 0.9pt}C{0.70cm}!{\vrule width 0.9pt}C{1.12cm}!{\vrule width 0.9pt}C{1.15cm}|C{1.15cm}|C{1.15cm}|C{1.15cm}!{\vrule width 0.9pt}C{1.15cm}|C{1.15cm}|C{1.15cm}|C{1.15cm}!{\vrule width 0.9pt}C{1.15cm}|C{1.15cm}|C{1.15cm}|C{1.15cm}!{\vrule width 0.9pt}C{1.15cm}|C{1.15cm}|C{1.15cm}|C{1.15cm}!{\vrule width 0.9pt}C{1.15cm}|C{1.15cm}|C{1.15cm}|C{1.15cm}!{\vrule width 0.9pt}C{1.15cm}|C{1.15cm}|C{1.15cm}|C{1.15cm}!{\vrule width 0.9pt}C{1.15cm}|C{1.15cm}|C{1.15cm}|C{1.15cm}!{\vrule width 0.9pt}}
    \thickhline
    \multirow[c]{2}{*}{QP} & \multirow[c]{2}{*}{Class} & \multirow[c]{2}{*}{\scalebox{0.92}{Sequence}} & \multicolumn{4}{c!{\vrule width 0.9pt}}{{\footnotesize\renewcommand{\arraystretch}{1.0}\begin{tabular}{@{}c@{}}\rule{0pt}{2.4ex}STDR\\[0.25ex](TMM 2023)\end{tabular}}} & \multicolumn{4}{c!{\vrule width 0.9pt}}{{\footnotesize\renewcommand{\arraystretch}{1.0}\begin{tabular}{@{}c@{}}\rule{0pt}{2.4ex}STFF\\[0.25ex](TBC 2025)\end{tabular}}} & \multicolumn{4}{c!{\vrule width 0.9pt}}{{\footnotesize\renewcommand{\arraystretch}{1.0}\begin{tabular}{@{}c@{}}\rule{0pt}{2.4ex}MW-GAN+\\[0.25ex](TCSVT 2021)\end{tabular}}} & \multicolumn{4}{c!{\vrule width 0.9pt}}{{\footnotesize\renewcommand{\arraystretch}{1.0}\begin{tabular}{@{}c@{}}\rule{0pt}{2.4ex}HFGAT\\[0.25ex](AAAI 2026)\end{tabular}}} & \multicolumn{4}{c!{\vrule width 0.9pt}}{{\footnotesize\renewcommand{\arraystretch}{1.0}\begin{tabular}{@{}c@{}}\rule{0pt}{2.4ex}SeedVR2-3B\\[0.25ex](ICLR 2026)\end{tabular}}} & \multicolumn{4}{c!{\vrule width 0.9pt}}{{\footnotesize\renewcommand{\arraystretch}{1.0}\begin{tabular}{@{}c@{}}\rule{0pt}{2.4ex}DiffCVE-50\\[0.25ex](Ours)\end{tabular}}} & \multicolumn{4}{c!{\vrule width 0.9pt}}{{\footnotesize\renewcommand{\arraystretch}{1.0}\begin{tabular}{@{}c@{}}\rule{0pt}{2.4ex}DiffCVE-1\\[0.25ex](Ours)\end{tabular}}} \\
    \thickcline{4-31}
    & & & {\footnotesize MUSIQ} & {\footnotesize CLIPIQA} & {\footnotesize DOVER} & {\footnotesize NIQE} & {\footnotesize MUSIQ} & {\footnotesize CLIPIQA} & {\footnotesize DOVER} & {\footnotesize NIQE} & {\footnotesize MUSIQ} & {\footnotesize CLIPIQA} & {\footnotesize DOVER} & {\footnotesize NIQE} & {\footnotesize MUSIQ} & {\footnotesize CLIPIQA} & {\footnotesize DOVER} & {\footnotesize NIQE} & {\footnotesize MUSIQ} & {\footnotesize CLIPIQA} & {\footnotesize DOVER} & {\footnotesize NIQE} & {\footnotesize MUSIQ} & {\footnotesize CLIPIQA} & {\footnotesize DOVER} & {\footnotesize NIQE} & {\footnotesize MUSIQ} & {\footnotesize CLIPIQA} & {\footnotesize DOVER} & {\footnotesize NIQE} \\
    \thickhline
    \footnotesize
    \multirow[c]{19}{*}{42} & \multirow[c]{2}{*}{A} & 1 & -10.603 & 0.096 & 7.085 & 1.069 & -10.217 & 0.113 & 9.447 & 1.152 & -2.374 & 0.152 & 10.995 & -0.047 & -7.580 & 0.131 & \second{20.194} & -0.253 & \best{15.823} & 0.177 & 12.797 & -1.307 & 5.687 & \second{0.249} & 17.811 & \second{-1.499} & \second{14.686} & \best{0.263} & \best{20.383} & \best{-2.229} \\
    \cline{3-31}
     &  & 2 & 5.364 & 0.041 & 10.565 & 0.706 & 5.972 & 0.045 & 9.534 & 0.813 & 10.048 & 0.078 & 12.061 & -0.880 & 10.495 & 0.107 & 15.913 & -0.596 & \best{20.327} & 0.115 & 15.990 & -1.777 & 9.471 & \best{0.132} & \second{18.359} & \best{-2.105} & \second{14.666} & \second{0.117} & \best{20.209} & \second{-2.073} \\
    \thickcline{2-31}
     & \multirow[c]{5}{*}{B} & 3 & 10.166 & 0.045 & 13.467 & 0.454 & \best{11.253} & 0.041 & 14.392 & 0.550 & 0.148 & 0.175 & 16.679 & 0.006 & 3.137 & \best{0.252} & \best{25.822} & -0.487 & \second{10.737} & 0.156 & 15.841 & -0.522 & 5.370 & 0.230 & 14.342 & \second{-1.532} & 7.735 & \second{0.242} & \second{17.023} & \best{-1.642} \\
    \cline{3-31}
     &  & 4 & 8.527 & 0.027 & 9.139 & 0.774 & 8.889 & 0.031 & 11.211 & 0.858 & 8.719 & 0.073 & 20.027 & -0.986 & 8.768 & 0.091 & \best{25.541} & -1.109 & 12.671 & 0.039 & 18.112 & -0.909 & \second{14.527} & \second{0.123} & 23.781 & \second{-1.820} & \best{16.980} & \best{0.170} & \second{24.387} & \best{-2.170} \\
    \cline{3-31}
     &  & 5 & 15.167 & 0.062 & 4.852 & 0.205 & 15.699 & 0.060 & 3.705 & 0.378 & 17.617 & 0.080 & 20.704 & -1.097 & 19.352 & 0.152 & \best{26.425} & -1.455 & \second{27.819} & \second{0.366} & 21.576 & -1.545 & 24.201 & 0.311 & 24.479 & \second{-2.533} & \best{30.144} & \best{0.410} & \second{26.131} & \best{-2.606} \\
    \cline{3-31}
     &  & 6 & 7.230 & 0.051 & 19.524 & 0.411 & 6.699 & 0.052 & 16.614 & 0.607 & 10.822 & 0.098 & 32.496 & -1.690 & 11.038 & 0.136 & \second{40.315} & -1.881 & \second{16.055} & 0.096 & 33.508 & \second{-2.290} & 13.907 & \second{0.231} & 39.786 & \best{-2.311} & \best{18.472} & \best{0.315} & \best{45.145} & -2.247 \\
    \cline{3-31}
     &  & 7 & 6.665 & 0.047 & 17.311 & 0.301 & 7.405 & 0.041 & 17.422 & 0.401 & 6.916 & 0.110 & 36.546 & -0.738 & 11.008 & 0.140 & \second{45.661} & -1.893 & \second{18.753} & 0.108 & 37.198 & -1.849 & 16.269 & \second{0.225} & 42.199 & \best{-3.573} & \best{20.230} & \best{0.226} & \best{47.525} & \second{-3.148} \\
    \thickcline{2-31}
     & \multirow[c]{4}{*}{C} & 8 & 10.265 & 0.046 & 19.320 & 0.467 & 10.869 & 0.042 & 17.405 & 0.629 & 8.760 & 0.055 & 30.596 & -0.209 & 11.436 & 0.025 & \second{35.113} & -0.891 & \best{20.090} & 0.045 & \best{36.966} & -0.928 & 13.543 & \second{0.121} & 33.578 & \best{-1.663} & \second{17.229} & \best{0.135} & 34.850 & \second{-1.590} \\
    \cline{3-31}
     &  & 9 & 11.885 & 0.035 & 23.449 & 0.262 & 12.114 & 0.040 & 23.261 & 0.485 & 11.351 & -0.021 & 31.301 & -0.104 & 12.837 & 0.033 & 37.057 & -1.119 & \second{17.792} & 0.038 & 34.753 & -1.295 & 14.350 & \second{0.064} & \best{42.734} & \best{-2.173} & \best{20.548} & \best{0.092} & \second{42.345} & \second{-1.957} \\
    \cline{3-31}
     &  & 10 & 9.262 & 0.089 & 13.371 & -0.199 & 9.881 & 0.075 & 13.024 & 0.155 & 5.682 & 0.036 & 21.923 & 1.248 & 9.178 & 0.099 & \second{29.126} & -0.592 & \second{12.909} & 0.202 & 23.634 & -0.706 & 12.472 & \second{0.318} & 28.615 & \second{-2.064} & \best{15.844} & \best{0.347} & \best{30.409} & \best{-2.081} \\
    \cline{3-31}
     &  & 11 & 9.772 & 0.039 & 11.902 & 0.536 & 9.946 & 0.037 & 12.739 & 0.751 & 10.674 & 0.161 & 21.771 & -0.268 & 12.159 & 0.218 & \best{26.813} & -1.602 & 12.953 & 0.271 & 16.221 & -1.130 & \second{16.432} & \best{0.281} & 24.511 & \second{-2.641} & \best{19.283} & \second{0.274} & \second{26.280} & \best{-3.047} \\
    \thickcline{2-31}
     & \multirow[c]{4}{*}{D} & 12 & 7.346 & 0.027 & 11.688 & 0.593 & 8.113 & 0.028 & 13.972 & 1.243 & -0.371 & 0.123 & 5.931 & 1.067 & 6.785 & 0.220 & 16.888 & -0.571 & 6.886 & \second{0.246} & 19.153 & 0.272 & \second{10.520} & 0.231 & \second{22.706} & \second{-2.105} & \best{12.594} & \best{0.252} & \best{26.195} & \best{-2.610} \\
    \cline{3-31}
     &  & 13 & 6.666 & 0.051 & 18.116 & -0.104 & 6.816 & 0.048 & 17.718 & 0.275 & 7.316 & 0.040 & 14.190 & -1.046 & 8.299 & 0.025 & 28.626 & -2.354 & \best{17.442} & 0.107 & \second{42.192} & -2.182 & \second{13.580} & \second{0.116} & \best{44.421} & \best{-4.062} & 13.423 & \best{0.136} & 37.198 & \second{-3.526} \\
    \cline{3-31}
     &  & 14 & 7.664 & 0.099 & 18.453 & -0.373 & 8.441 & 0.055 & 20.470 & 0.229 & 5.523 & 0.076 & 18.615 & 0.116 & 9.361 & 0.153 & 32.904 & -2.021 & 16.108 & 0.253 & 43.791 & -1.571 & \second{21.321} & \best{0.414} & \best{56.595} & \best{-3.452} & \best{21.558} & \second{0.413} & \second{53.632} & \second{-3.316} \\
    \cline{3-31}
     &  & 15 & 10.607 & 0.069 & 18.855 & 0.466 & 10.790 & 0.070 & 17.769 & 0.874 & 11.015 & 0.206 & 17.158 & -0.207 & 11.289 & 0.218 & 32.739 & -1.367 & \best{21.116} & \best{0.306} & 41.392 & -1.401 & 18.962 & 0.243 & \best{43.352} & \second{-2.989} & \second{20.559} & \second{0.261} & \second{42.949} & \best{-3.253} \\
    \thickcline{2-31}
     & \multirow[c]{3}{*}{E} & 16 & 13.968 & 0.057 & 18.877 & 0.133 & 13.357 & 0.059 & 17.591 & 0.241 & 19.452 & 0.048 & 40.124 & -1.064 & 21.256 & 0.139 & 44.991 & -1.412 & \second{32.899} & \second{0.254} & \second{58.360} & -2.105 & 31.763 & 0.246 & 57.625 & \second{-2.443} & \best{35.195} & \best{0.290} & \best{61.519} & \best{-2.515} \\
    \cline{3-31}
     &  & 17 & 10.995 & 0.018 & 17.639 & 0.475 & 10.990 & 0.024 & 16.407 & 0.397 & 16.480 & 0.144 & 27.588 & -1.249 & 21.126 & 0.153 & 35.519 & -1.549 & 31.937 & 0.246 & 43.618 & -1.614 & \second{32.165} & \second{0.282} & \second{46.499} & \second{-2.327} & \best{34.846} & \best{0.337} & \best{49.094} & \best{-3.096} \\
    \cline{3-31}
     &  & 18 & 12.948 & 0.074 & 27.652 & 0.160 & 13.131 & 0.068 & 27.877 & 0.412 & 14.421 & 0.166 & 44.291 & -1.246 & 17.730 & 0.246 & 50.995 & -1.851 & 26.194 & 0.245 & 52.749 & -1.824 & \second{30.545} & \second{0.373} & \second{57.136} & \best{-2.850} & \best{30.974} & \best{0.378} & \best{58.094} & \second{-2.663} \\
    \thickcline{2-31}
     & \multicolumn{2}{c!{\vrule width 0.9pt}}{Avg.} & 8.549 & 0.054 & 15.626 & 0.352 & 8.897 & 0.052 & 15.587 & 0.581 & 9.011 & 0.100 & 23.500 & -0.466 & 10.982 & 0.141 & 31.702 & -1.277 & \second{18.806} & 0.182 & 31.547 & -1.371 & 16.949 & \second{0.233} & \second{35.474} & \second{-2.452} & \best{20.276} & \best{0.259} & \best{36.854} & \best{-2.543} \\
    \thickhline
    \multirow[c]{1}{*}{37} & \multicolumn{2}{c!{\vrule width 0.9pt}}{Avg.} & 5.921 & 0.045 & 8.357 & 0.313 & 6.381 & 0.046 & 8.238 & 0.378 & 7.016 & 0.065 & 12.187 & -1.407 & 7.989 & 0.109 & 17.102 & -0.823 & \best{13.933} & \best{0.182} & 16.271 & -0.894 & 10.265 & \second{0.175} & \second{18.032} & \best{-1.721} & \second{11.916} & \best{0.182} & \best{18.572} & \second{-1.656} \\
    \thickhline
    \multirow[c]{1}{*}{32} & \multicolumn{2}{c!{\vrule width 0.9pt}}{Avg.} & 3.949 & 0.031 & 4.431 & 0.301 & 3.999 & 0.028 & 4.713 & 0.331 & 5.067 & 0.045 & 7.735 & \second{-1.122} & 5.310 & 0.059 & 8.895 & -0.560 & \best{9.922} & \best{0.146} & 8.471 & -0.568 & 7.109 & 0.122 & \second{10.211} & \best{-1.221} & \second{8.046} & \second{0.124} & \best{10.347} & -1.102 \\
    \thickhline
    \end{tabular}}
    \par\vspace*{3pt}
    \noindent\begin{minipage}{\textwidth}
    \raggedright
    \scriptsize
    The resolution of class A is $2560\times1600$, 1: PeopleOnStreet; 2: Traffic.\\
    The resolution of class B is $1920\times1080$, 3: BQTerrace; 4: BasketballDrive; 5: Cactus; 6: Kimono1; 7: ParkScene.\\
    The resolution of class C is $832\times480$, 8: BQMall; 9: BasketballDrill; 10: PartyScene; 11: RaceHorses.\\
    The resolution of class D is $416\times240$, 12: BQSquare; 13: BasketballPass; 14: BlowingBubbles; 15: RaceHorses.\\
    The resolution of class E is $1280\times720$, 16: FourPeople; 17: Johnny; 18: KristenAndSara.\\
    \end{minipage}
\end{table*}

\begin{figure*}[!t]
    \centering
    \includegraphics[width=1.0\textwidth]{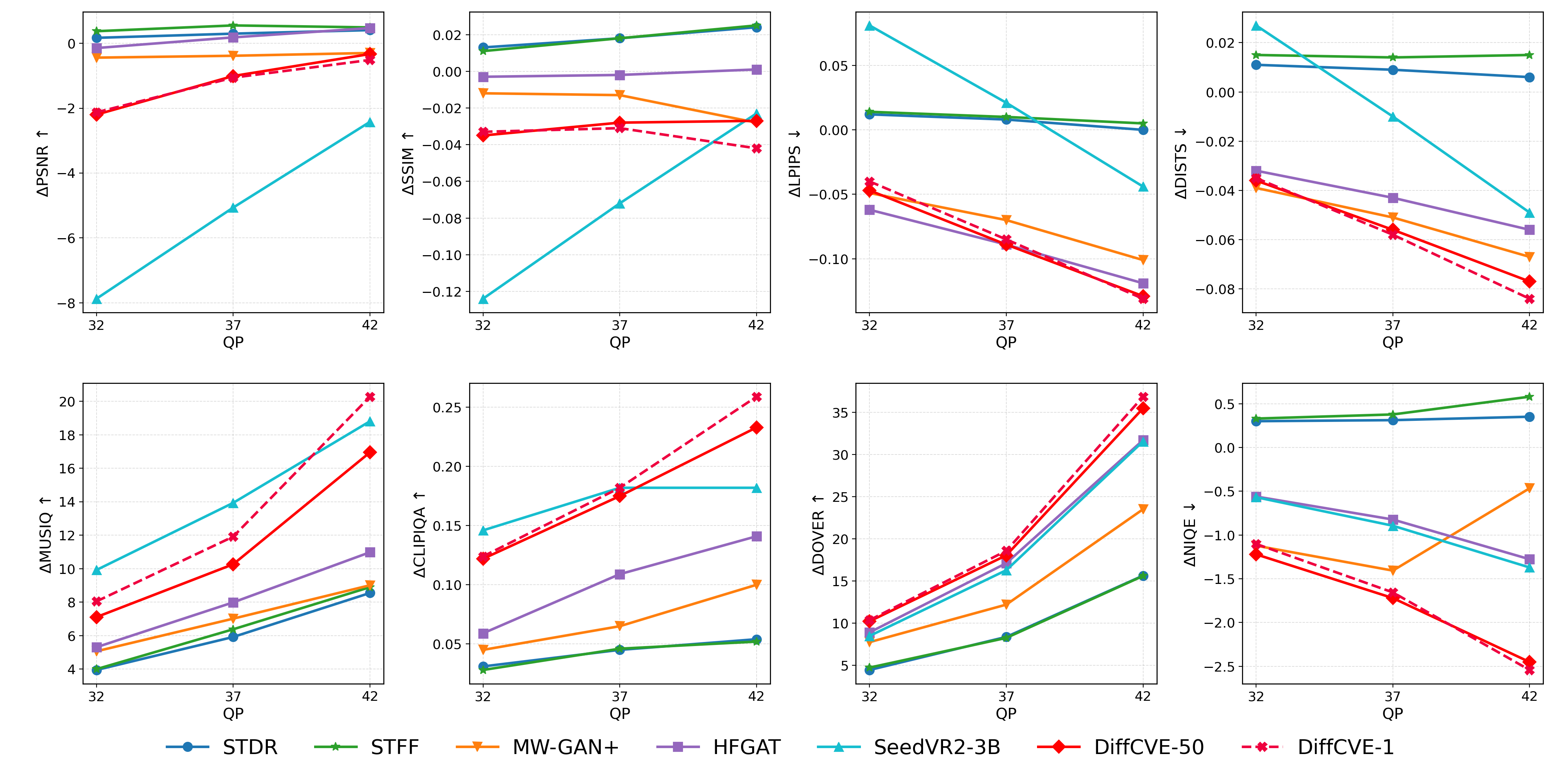}
    \caption{Overall trends of the eight evaluation metrics across QP levels. The first row shows full-reference metrics and the second row shows no-reference metrics.}
    \label{fig:metric_curves}
    \vspace{-1pt}
\end{figure*}

\subsection{Comparison with the State-of-the-Art Methods}
The proposed method is compared with five representative baselines from three categories. STDR~\cite{luo2022spatio} and STFF~\cite{wang2025stff} are distortion-oriented compressed video quality enhancement methods. MW-GAN+~\cite{wang2021mw} and HFGAT~\cite{peng2026hierarchical} are compressed video perceptual quality enhancement methods. The most closely related diffusion-based compressed video enhancement methods, LViCAR~\cite{gehlot2025lvicar} and SDATC~\cite{an2025spatial}, are not included because their code is not publicly available and the experimental settings reported in their papers differ from ours. Consequently, a direct comparison with their reported results is not feasible. We therefore select SeedVR2-3B~\cite{wang2025seedvr2}, a recent generic video restoration method, as the diffusion baseline. STDR, STFF, MW-GAN+, and HFGAT are retrained under the same compression settings and training data, while SeedVR2-3B is reproduced based on its public implementation.

\textbf{Quantitative Results:} Table~\ref{tab:compare_fr} reports full-reference results. STFF obtains the best average $\Delta$PSNR and $\Delta$SSIM at QP 42 and 37, showing the advantage of distortion-oriented methods in fidelity reconstruction. By contrast, DiffCVE-50 and DiffCVE-1 achieve the best or second-best average $\Delta$LPIPS and $\Delta$DISTS at QP 42 and 37, indicating stronger perceptual restoration under heavy compression. At QP 32, the compressed inputs already preserve more visual information and contain fewer severe artifacts, leaving less generative restoration space for diffusion-based enhancement. Therefore, MW-GAN+ and HFGAT are slightly better on average $\Delta$LPIPS and $\Delta$DISTS, while the proposed method remains competitive, especially on $\Delta$DISTS.

To further evaluate perceptual quality, Table~\ref{tab:compare_nr} gives no-reference results. At QP 42, DiffCVE-1 ranks first on all four average metrics, and DiffCVE-50 ranks second on $\Delta$CLIPIQA, $\Delta$DOVER, and $\Delta$NIQE. At QP 37, DiffCVE-1 leads in $\Delta$CLIPIQA and $\Delta$DOVER, while DiffCVE-50 achieves the best $\Delta$NIQE. At QP 32, SeedVR2-3B is more competitive on $\Delta$MUSIQ and $\Delta$CLIPIQA due to its strong generic diffusion prior, but DiffCVE-1 still obtains the best $\Delta$DOVER and DiffCVE-50 the best $\Delta$NIQE. These results indicate that the proposed method provides clear perceptual benefits under stronger compression and remains competitive as compression becomes milder.

Fig.~\ref{fig:metric_curves} further visualizes the metric trends across QP levels. The curves are consistent with the table results: as QP increases and the bitrate decreases, the performance gap among different methods becomes larger, suggesting that stronger compression makes perceptual enhancement more challenging and also leaves more room for generative restoration. On fidelity-oriented metrics, STFF and STDR maintain clearer advantages in $\Delta$PSNR and $\Delta$SSIM. However, on perceptual metrics, the proposed DiffCVE variants show stronger overall performance across different compression levels. DiffCVE-50 and DiffCVE-1 achieve more favorable trends on $\Delta$LPIPS and $\Delta$DISTS, and they also provide the leading or near-leading performance on $\Delta$MUSIQ, $\Delta$CLIPIQA, $\Delta$DOVER, and $\Delta$NIQE under stronger compression. In summary, these trends further demonstrate the effectiveness of the proposed method in improving perceptual quality under compressed video settings.

\begin{table}[!t]
    \caption{Perceptual quality comparison on the LDV Track 1 test set. Each row corresponds to one method, and each QP group reports average $\Delta$LPIPS$\downarrow$ and $\Delta$DISTS$\downarrow$. Best and second-best values are marked in boldface and underlined, respectively.}
    \label{tab:ldv}
    \centering
    \footnotesize
    \renewcommand{\arraystretch}{1.3}
    \setlength{\tabcolsep}{4pt}
    \resizebox{\columnwidth}{!}{%
    \begin{tabular}{c cc cc cc}
    \toprule
    \multirow[c]{2}{*}{Method} & \multicolumn{2}{c}{QP 42} & \multicolumn{2}{c}{QP 37} & \multicolumn{2}{c}{QP 32} \\
    & LPIPS & DISTS & LPIPS & DISTS & LPIPS & DISTS \\
    \midrule
    HFGAT~\cite{peng2026hierarchical} & -0.106 & -0.050 & \second{-0.082} & -0.037 & \best{-0.055} & \second{-0.026} \\
    MGLD-VSR~\cite{yang2024motion} & -0.050 & -0.053 & 0.004 & -0.016 & 0.061 & 0.023 \\
    DLoRAL~\cite{sun2026one} & -0.065 & -0.064 & 0.005 & -0.013 & 0.075 & 0.034 \\
    SeedVR2-3B~\cite{wang2025seedvr2} & -0.086 & \second{-0.076} & -0.017 & -0.026 & 0.057 & 0.028 \\
    DiffCVE-50 (Ours) & \best{-0.126} & -0.073 & \best{-0.084} & \second{-0.045} & \second{-0.034} & -0.020 \\
    DiffCVE-1 (Ours) & \second{-0.119} & \best{-0.091} & -0.077 & \best{-0.058} & -0.031 & \best{-0.028} \\
    \bottomrule
    \end{tabular}}
\end{table}

\begin{figure*}[!t]
    \centering
    \includegraphics[width=1.0\textwidth]{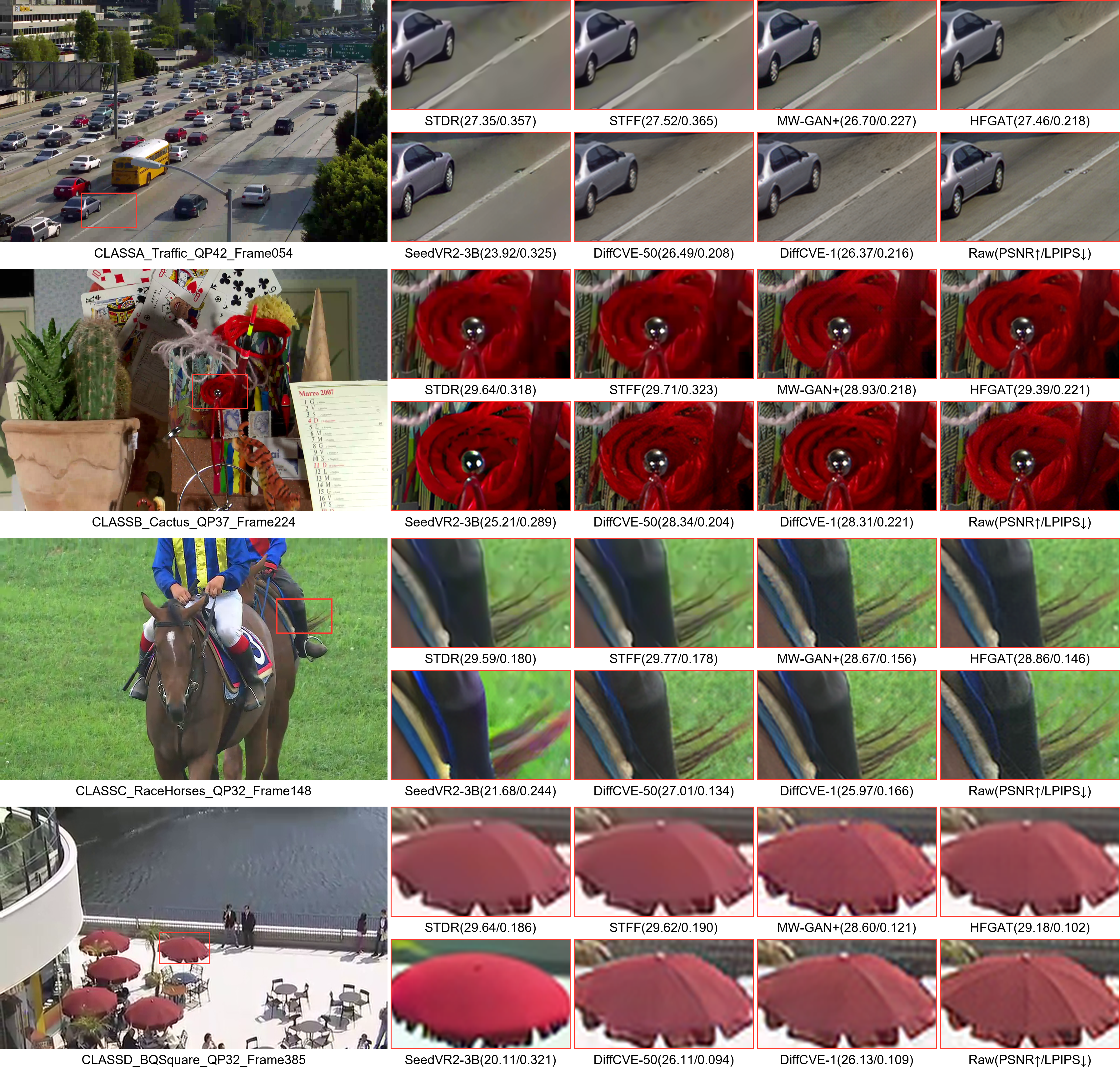}
    \caption{Subjective visual comparison with PSNR$\uparrow$ / LPIPS$\downarrow$ reported for each enhanced result. The proposed variants recover more perceptually faithful details despite lower PSNR than several competing methods.}
    \label{fig:visual_compare}
    \vspace{-1pt}
\end{figure*}

To further evaluate the robustness of our proposed variants, we test them on the LDV Track 1 test set~\cite{yang2022ntire}. The performance is compared with HFGAT and SeedVR2-3B, which show the strongest perceptual performance among the compared frameworks in our main experiments. Additionally, we include two diffusion-based video enhancement methods, MGLD-VSR~\cite{yang2024motion} and DLoRAL~\cite{sun2026one}, in the comparison. Table~\ref{tab:ldv} summarizes the average $\Delta$LPIPS and $\Delta$DISTS results under three QP levels. Overall, the proposed variants achieve competitive or superior perceptual quality compared with the evaluated methods, further demonstrating their effectiveness under diverse video compression scenarios.

\begin{table}[!t]
    \caption{Perceptual quality comparison on H.265 compressed JCT-VC Class C and E sequences. Best and second-best values for each QP and metric are marked in bold and underlined, respectively.}
    \label{tab:h265_classce}
    \centering
    \footnotesize
    \renewcommand{\arraystretch}{1.15}
    \setlength{\tabcolsep}{2pt}
    \resizebox{\columnwidth}{!}{\begin{tabular}{c l c c c c}
    \toprule
    QP & Method & $\Delta$LPIPS$\downarrow$ & $\Delta$DISTS$\downarrow$ & $\Delta$MUSIQ$\uparrow$ & $\Delta$NIQE$\downarrow$ \\
    \midrule
    42 & HFGAT~\cite{peng2026hierarchical} & \second{-0.060} & -0.031 & 13.557 & -0.272 \\
    42 & SeedVR2-3B~\cite{wang2025seedvr2} & 0.007 & -0.018 & \second{17.357} & -0.129 \\
    42 & DiffCVE-50 (Ours) & -0.048 & \second{-0.033} & 15.140 & \best{-1.201} \\
    42 & DiffCVE-1 (Ours) & \best{-0.070} & \best{-0.050} & \best{17.723} & \second{-1.073} \\
    \midrule
    37 & HFGAT~\cite{peng2026hierarchical} & \best{-0.046} & -0.024 & 8.850 & -0.240 \\
    37 & SeedVR2-3B~\cite{wang2025seedvr2} & 0.061 & 0.014 & \best{11.243} & -0.264 \\
    37 & DiffCVE-50 (Ours) & -0.043 & \second{-0.034} & 9.349 & \best{-1.013} \\
    37 & DiffCVE-1 (Ours) & \second{-0.044} & \best{-0.036} & \second{10.766} & \second{-0.853} \\
    \midrule
    32 & HFGAT~\cite{peng2026hierarchical} & \best{-0.027} & -0.013 & 5.763 & -0.296 \\
    32 & SeedVR2-3B~\cite{wang2025seedvr2} & 0.105 & 0.043 & \best{8.187} & -0.297 \\
    32 & DiffCVE-50 (Ours) & \second{-0.014} & \best{-0.020} & 6.900 & \best{-0.901} \\
    32 & DiffCVE-1 (Ours) & -0.008 & \second{-0.017} & \second{7.787} & \second{-0.787} \\
    \bottomrule
    \end{tabular}}
\end{table}

\rev{We further evaluate the proposed variants on H.265 compressed JCT-VC Class C and E sequences to assess their robustness across compression standards. Table~\ref{tab:h265_classce} compares HFGAT, SeedVR2-3B, DiffCVE-50, and DiffCVE-1 at QP 32, 37, and 42. Across all three QPs, our variants achieve the best or near-best results on all four metrics. In particular, the two DiffCVE variants achieve the best and second-best $\Delta$DISTS and $\Delta$NIQE at all three QPs. These results further demonstrate the robustness of the proposed variants in perceptual enhancement across compression standards.}

\begin{table}[!t]
    \caption{Temporal consistency comparison measured by $\Delta$tOF~\cite{chu2020learning}. QP 42 sequence indices follow Table~\ref{tab:compare_fr}; TI denotes mean temporal information~\cite{itut2023p910}. Average results over 18 sequences are given for QP 42, 37, and 32. Lower $\Delta$tOF is better. Best and second-best values in each row are marked in bold and underlined, respectively.}
    \label{tab:temporal_tof}
    \centering
    \scriptsize
    \renewcommand{\arraystretch}{1.15}
    \setlength{\tabcolsep}{1pt}
    \resizebox{\columnwidth}{!}{%
    \begin{tabular}{c c c@{\hspace{4pt}} c c c c c c c}
    \toprule
    \multirow{2}{*}{QP} & \multirow{2}{*}{Seq.} & \multirow{2}{*}{TI} & \multicolumn{7}{c}{$\Delta$tOF $\downarrow$} \\
    \cmidrule(lr){4-10}
    & & & STDR & STFF & MW-GAN+ & HFGAT & SeedVR2-3B & DiffCVE-50 & DiffCVE-1 \\
    \midrule
    \multirow{18}{*}{42} & 1 & 22.14 & -0.0097 & 0.0188 & \second{-0.0464} & \best{-0.0737} & 0.4217 & 0.0277 & 0.0601 \\
     & 2 & 11.78 & -0.0451 & \second{-0.0564} & -0.0549 & \best{-0.0736} & 0.1796 & 0.0475 & 0.0483 \\
     & 3 & 17.76 & 0.0195 & 0.0104 & 0.0111 & 0.0099 & 0.1363 & \second{-0.0310} & \best{-0.0494} \\
     & 4 & 19.90 & 0.2290 & 0.2564 & -0.3092 & -0.2356 & -0.3911 & \second{-0.3939} & \best{-0.5182} \\
     & 5 & 13.14 & -0.0022 & 0.0131 & -0.0493 & -0.0718 & -0.1055 & \second{-0.1206} & \best{-0.1777} \\
     & 6 & 11.88 & -0.0780 & -0.0579 & -0.2455 & -0.2420 & \second{-0.4126} & -0.3910 & \best{-0.4393} \\
     & 7 & 11.78 & -0.0725 & -0.0871 & -0.0533 & \second{-0.1134} & -0.0882 & \best{-0.1302} & -0.0839 \\
     & 8 & 21.92 & -0.0053 & 0.0079 & -0.0535 & -0.0600 & 0.0569 & \second{-0.2446} & \best{-0.2545} \\
     & 9 & 17.48 & 0.0722 & 0.1034 & \best{-0.0605} & \second{-0.0497} & 0.0424 & -0.0044 & 0.0126 \\
     & 10 & 12.90 & \best{-0.0993} & -0.0889 & -0.0342 & \second{-0.0948} & 0.1122 & -0.0787 & -0.0838 \\
     & 11 & 28.14 & 0.1753 & 0.2483 & -0.2764 & -0.2276 & -0.1748 & \second{-0.4121} & \best{-0.4823} \\
     & 12 & 13.64 & -0.0522 & -0.0016 & -0.0164 & \best{-0.0642} & 0.0045 & \second{-0.0601} & -0.0267 \\
     & 13 & 19.19 & 0.0287 & 0.0609 & -0.1018 & -0.0870 & -0.0400 & \second{-0.1491} & \best{-0.2478} \\
     & 14 & 9.50 & \second{-0.1070} & \best{-0.1103} & -0.0476 & -0.1018 & -0.0041 & -0.0787 & -0.0802 \\
     & 15 & 25.47 & 0.0190 & 0.0537 & -0.0919 & -0.0999 & -0.0979 & \second{-0.1743} & \best{-0.2193} \\
     & 16 & 4.35 & \second{-0.0209} & -0.0195 & -0.0151 & \best{-0.0244} & 0.0284 & -0.0051 & -0.0101 \\
     & 17 & 2.95 & \second{-0.0137} & -0.0058 & -0.0107 & \best{-0.0165} & 0.0347 & 0.0469 & 0.0482 \\
     & 18 & 4.15 & -0.0157 & \second{-0.0225} & -0.0147 & \best{-0.0249} & 0.0517 & 0.0094 & 0.0252 \\
    \midrule
    42 & Avg. & -- & 0.0012 & 0.0179 & -0.0817 & -0.0917 & -0.0137 & \second{-0.1190} & \best{-0.1377} \\
    37 & Avg. & -- & -0.0082 & -0.0009 & -0.0671 & \second{-0.0859} & 0.1353 & -0.0799 & \best{-0.0964} \\
    32 & Avg. & -- & -0.0138 & 0.0003 & \second{-0.0652} & \best{-0.0771} & 0.2935 & -0.0160 & -0.0200 \\
    \bottomrule
    \end{tabular}}
\end{table}

\begin{figure*}[!t]
    \centering
    \includegraphics[width=1.0\textwidth]{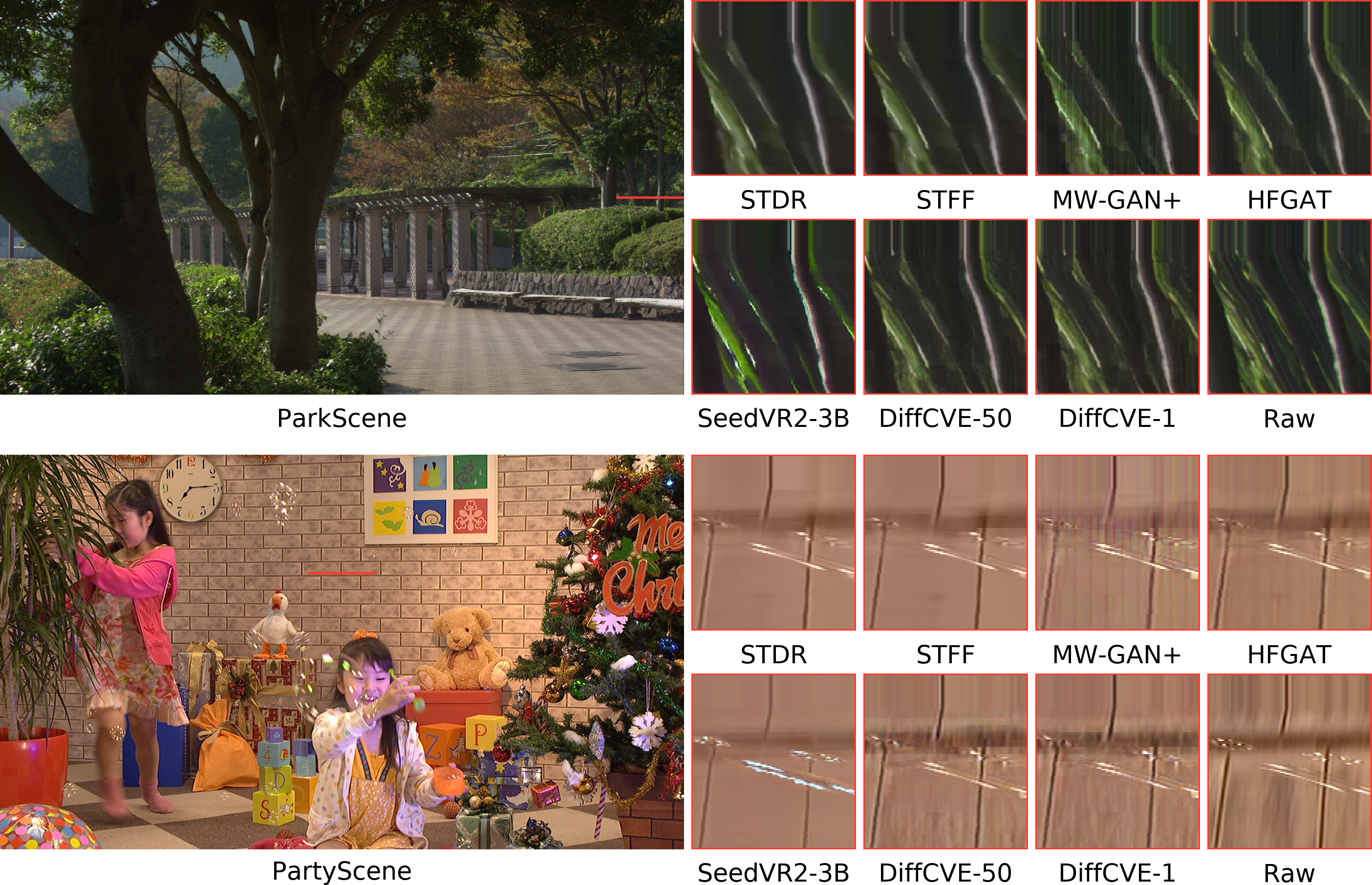}
    \caption{Temporal consistency visualization under QP 42. Consecutive frames of each method are aligned in the same row for direct comparison of temporal stability and texture fidelity. The stacked strip on the right is formed by concatenating the pixels sampled along the red line across successive frames.}
    \label{fig:temporal_stack}
\end{figure*}

\textbf{Qualitative Results:} \rev{The visual comparisons in Fig.~\ref{fig:visual_compare} illustrate the perceptual benefits of the proposed variants at different QP levels. Although STDR and STFF achieve higher PSNR in these examples, the proposed variants yield lower LPIPS and recover more natural textures and structures. According to the perception-distortion trade-off theory~\cite{blau2018perception}, optimal reconstruction fidelity and optimal perceptual quality cannot generally be achieved simultaneously. Our method therefore prioritizes perceptual quality.} In the first row, the \emph{Traffic} example shows that the proposed method generates richer ground textures, whereas the competing methods produce relatively blurred surfaces. In the second row, the \emph{Cactus} example shows clearer rope-like textures and local structures restored by the proposed method, while the other methods tend to oversmooth these details. \rev{In the third row, the \emph{RaceHorses} example shows finer and more natural texture details restored by the proposed variants.} \rev{In the fourth row, the \emph{BQSquare} example shows a more realistic parasol structure, with its shape and contours closer to the uncompressed reference.} \rev{These comparisons show that higher PSNR alone does not ensure better visual quality and demonstrate the perceptual benefit of our method.}

\begin{figure}[!t]
    \centering
    \includegraphics[width=0.92\columnwidth]{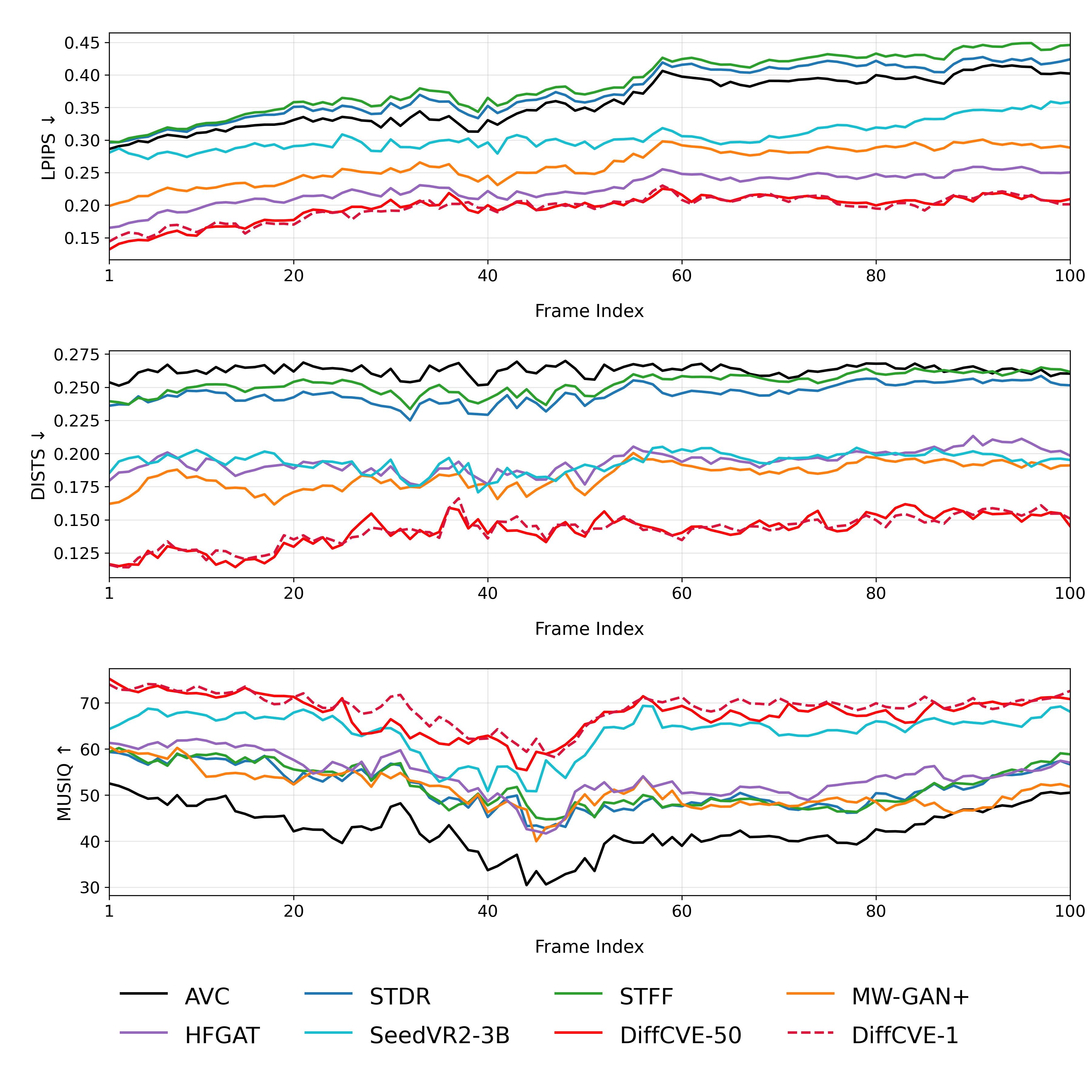}
    \caption{Frame-wise perceptual metrics on Class D \emph{BlowingBubbles} under QP 42.}
    \label{fig:blowingbubbles_curve}
\end{figure}

\textbf{Temporal Consistency:} Table~\ref{tab:temporal_tof} reports the $\Delta$tOF results for temporal consistency evaluation. \rev{Our method achieves the best average $\Delta$tOF values at QP 42 and 37.} At QP 32, HFGAT achieves the best value, whereas both proposed variants still remain competitive. \rev{We further investigate the temporal consistency of our method across different levels of temporal information (TI)~\cite{itut2023p910}. Higher TI indicates greater frame-to-frame variation and thus stronger temporal activity. All TI values are computed on the original uncompressed videos and averaged over time for each sequence. At QP 42, both DiffCVE variants improve temporal consistency on most sequences with higher TI, while improvements on lower-TI sequences are more limited and depend on the sequence. These results demonstrate the effectiveness of the proposed design in improving temporal consistency, particularly under heavier compression and for sequences with higher temporal activity.} Fig.~\ref{fig:blowingbubbles_curve} further presents frame-wise quality curves on \emph{BlowingBubbles}, where the proposed variants maintain lower LPIPS and DISTS values and higher MUSIQ values over the shown frames, suggesting that the perceptual improvement is preserved across consecutive frames. Fig.~\ref{fig:temporal_stack} further visualizes temporal behavior on \emph{ParkScene} and \emph{PartyScene} under QP 42. In the stacked strips, several competing methods either oversmooth the sampled structures or introduce vertical banding and texture distortions. By contrast, the proposed variants reconstruct more accurate local structures and textures while preserving good temporal consistency.

\textbf{Model Complexity:} Table~\ref{tab:model_complexity} reports the model complexity comparison on Class C, where all runtimes are measured on a single NVIDIA RTX 3090 GPU and reported as the average inference time per frame. Compared with the generic diffusion-based baseline SeedVR2-3B, DiffCVE-50 achieves faster inference with a smaller parameter scale, and the single-step DiffCVE-1 further reduces the runtime to 0.329s per frame. Notably, DiffCVE-1 is also slightly faster than STFF, showing the efficiency advantage of the proposed design. Although DiffCVE-1 is slower than lightweight non-diffusion methods such as STDR, MW-GAN+, and HFGAT, it achieves better restoration performance under heavily compressed conditions, demonstrating the benefit of its additional computational cost.

\begin{table}[!t]
    \caption{Model complexity comparison on Class C. All runtimes are measured on a single NVIDIA RTX 3090 GPU and reported as the average inference time per frame.}
    \label{tab:model_complexity}
    \centering
    \footnotesize
    \renewcommand{\arraystretch}{1.2}
    \setlength{\tabcolsep}{2.6pt}
    \begin{tabular}{c c c c c}
    \toprule
    Model & Type & Step & Trainable / Total Params & Runtime \\
    \midrule
    STDR & Non-diffusion & -- & 1.34M / 1.34M & 0.061s \\
    STFF & Non-diffusion & -- & 4.45M / 4.45M & 0.337s \\
    MW-GAN+ & Non-diffusion & -- & 6.41M / 6.41M & 0.080s \\
    HFGAT & Non-diffusion & -- & 1.70M / 1.70M & 0.075s \\
    SeedVR2-3B & Diffusion & 1 & -- / 3.6B & 3.120s \\
    DiffCVE-50 & Diffusion & 50 & 261.78M / 1.6B & 2.955s \\
    DiffCVE-1 & Diffusion & 1 & 261.78M / 1.6B & 0.329s \\
    \bottomrule
    \end{tabular}
    \par\vspace*{2pt}
    \begin{minipage}{\columnwidth}
    \scriptsize
    \textit{Note:} ``--'' indicates that the corresponding item is not applicable or unavailable.
    \end{minipage}
\end{table}

\subsection{Ablation Study}

Incremental ablation experiments are performed on Class D sequences, as summarized in Table~\ref{tab:ablation_main}. Model A is the Stable Diffusion v2.1 baseline with middle-layer temporal blocks and a video conditioning branch. ``LoRA'' denotes lightweight U-Net fine-tuning, \rev{``QP Emb.'' denotes a learnable embedding of the numerical QP without textual conditioning, while }``Text'' denotes the textual prompt for degradation semantics in CDSP, and ``Res'' and ``MV'' denote using residual and motion-vector coding priors, respectively, as conditioning guidance for the diffusion U-Net. ``C.P. in CVCM'' denotes using coding priors to modulate video condition generation in CVCM. ``T.V.'' denotes temporal convolution together with VAE encoder feature modulation in the decoder, and ``T.V.P.'' denotes the proposed CPWF.

\begin{table}[!t]
    \caption{Incremental ablation study on Class D sequences.}
    \label{tab:ablation_main}
    \centering
    \footnotesize
    \renewcommand{\arraystretch}{1.3}
    \setlength{\tabcolsep}{1.2pt}
    \newcommand{\abcheck}{\checkmark}
    \newcommand{\abhead}[1]{\parbox[c][7ex][c]{\linewidth}{\centering #1}}
    \resizebox{\columnwidth}{!}{%
    \begin{tabular}{!{\vrule width 0.9pt}C{0.62cm}!{\vrule width 0.9pt}C{0.78cm}!{\vrule width 0.9pt}C{0.88cm}|C{0.90cm}|C{0.86cm}|C{0.82cm}|C{0.82cm}|C{1.12cm}!{\vrule width 0.9pt}C{0.86cm}|C{0.90cm}!{\vrule width 0.9pt}C{1.38cm}|C{1.38cm}!{\vrule width 0.9pt}}
    \thickhline
    \multirow[c]{2}{*}{QP} & \multirow[c]{2}{*}{Model} & \multicolumn{6}{>{\centering\arraybackslash}m{5.40cm}!{\vrule width 0.9pt}}{Diffusion U-Net} & \multicolumn{2}{>{\centering\arraybackslash}m{1.76cm}!{\vrule width 0.9pt}}{VAE Decoder} & \multicolumn{2}{>{\centering\arraybackslash}m{2.76cm}!{\vrule width 0.9pt}}{Metrics} \\
    \cline{3-12}
    & & \abhead{LoRA} & \abhead{QP\\Emb.} & \abhead{Text} & \abhead{Res} & \abhead{MV} & \abhead{C.P. in\\CVCM} & \abhead{T.V.} & \abhead{T.V.P.} & \abhead{$\Delta$LPIPS $\downarrow$} & \abhead{$\Delta$DISTS $\downarrow$} \\
    \thickhline
    \multirow[c]{9}{*}{42} & A &  &  &  &  &  &  &  &  & -0.060 & -0.072 \\
    \cline{2-12}
     & B & \abcheck &  &  &  &  &  &  &  & -0.069 & -0.080 \\
    \cline{2-12}
     & C & \abcheck & \abcheck &  &  &  &  &  &  & -0.072 & -0.082 \\
    \cline{2-12}
     & D & \abcheck &  & \abcheck &  &  &  &  &  & -0.071 & -0.083 \\
    \cline{2-12}
     & E & \abcheck &  & \abcheck & \abcheck &  & \abcheck &  &  & -0.074 & -0.084 \\
    \cline{2-12}
     & F & \abcheck &  & \abcheck & \abcheck & \abcheck &  &  &  & -0.075 & -0.084 \\
    \cline{2-12}
     & G & \abcheck &  & \abcheck & \abcheck & \abcheck & \abcheck &  &  & -0.080 & -0.087 \\
    \cline{2-12}
     & H & \abcheck &  & \abcheck & \abcheck & \abcheck & \abcheck & \abcheck &  & -0.121 & -0.100 \\
    \cline{2-12}
     & I & \abcheck &  & \abcheck & \abcheck & \abcheck & \abcheck &  & \abcheck & -0.125 & -0.101 \\
    \thickhline
    \multirow[c]{9}{*}{37} & A &  &  &  &  &  &  &  &  & -0.024 & -0.049 \\
    \cline{2-12}
     & B & \abcheck &  &  &  &  &  &  &  & -0.034 & -0.057 \\
    \cline{2-12}
     & C & \abcheck & \abcheck &  &  &  &  &  &  & -0.035 & -0.058 \\
    \cline{2-12}
     & D & \abcheck &  & \abcheck &  &  &  &  &  & -0.037 & -0.059 \\
    \cline{2-12}
     & E & \abcheck &  & \abcheck & \abcheck &  & \abcheck &  &  & -0.034 & -0.058 \\
    \cline{2-12}
     & F & \abcheck &  & \abcheck & \abcheck & \abcheck &  &  &  & -0.036 & -0.060 \\
    \cline{2-12}
     & G & \abcheck &  & \abcheck & \abcheck & \abcheck & \abcheck &  &  & -0.040 & -0.062 \\
    \cline{2-12}
     & H & \abcheck &  & \abcheck & \abcheck & \abcheck & \abcheck & \abcheck &  & -0.080 & -0.076 \\
    \cline{2-12}
     & I & \abcheck &  & \abcheck & \abcheck & \abcheck & \abcheck &  & \abcheck & -0.084 & -0.077 \\
    \thickhline
    \multirow[c]{9}{*}{32} & A &  &  &  &  &  &  &  &  & 0.020 & -0.019 \\
    \cline{2-12}
     & B & \abcheck &  &  &  &  &  &  &  & 0.011 & -0.026 \\
    \cline{2-12}
     & C & \abcheck & \abcheck &  &  &  &  &  &  & 0.011 & -0.027 \\
    \cline{2-12}
     & D & \abcheck &  & \abcheck &  &  &  &  &  & 0.009 & -0.028 \\
    \cline{2-12}
     & E & \abcheck &  & \abcheck & \abcheck &  & \abcheck &  &  & 0.012 & -0.026 \\
    \cline{2-12}
     & F & \abcheck &  & \abcheck & \abcheck & \abcheck &  &  &  & 0.009 & -0.028 \\
    \cline{2-12}
     & G & \abcheck &  & \abcheck & \abcheck & \abcheck & \abcheck &  &  & 0.006 & -0.029 \\
    \cline{2-12}
     & H & \abcheck &  & \abcheck & \abcheck & \abcheck & \abcheck & \abcheck &  & -0.029 & -0.048 \\
    \cline{2-12}
     & I & \abcheck &  & \abcheck & \abcheck & \abcheck & \abcheck &  & \abcheck & -0.035 & -0.051 \\
    \thickhline
    \end{tabular}}\\
    \vspace{2pt}
\end{table}

\begin{figure}[!t]
    \centering
    \includegraphics[width=1.0\columnwidth]{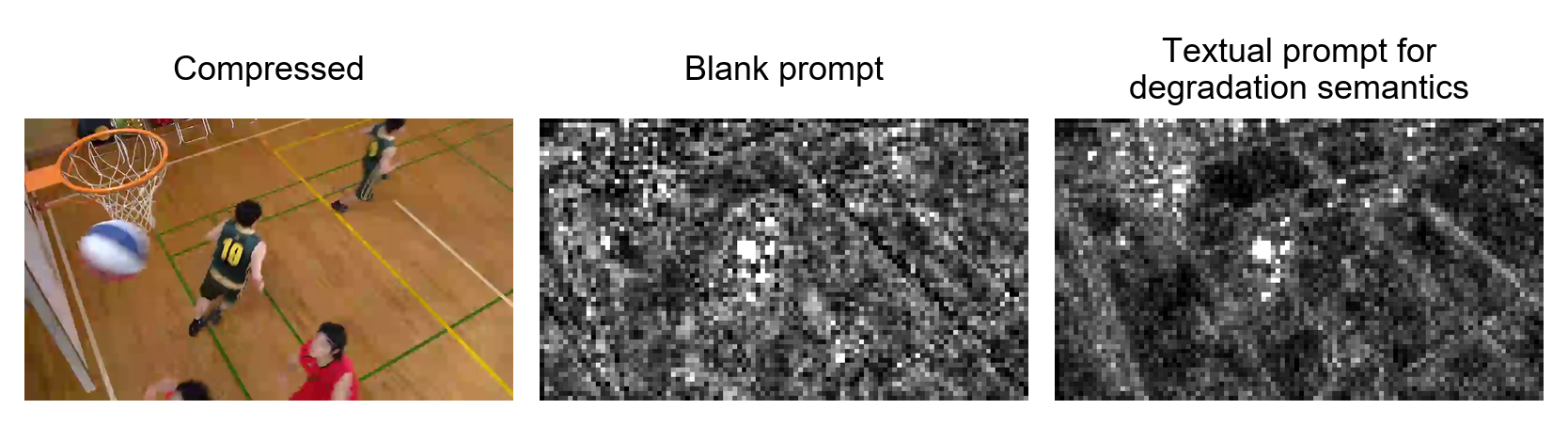}
\caption{Cross-attention visualization of CDSP. From left to right: compressed frame, cross-attention map with a blank prompt, and cross-attention map with the proposed textual prompt for degradation semantics.}
\label{fig:CDSP_prompt_vis}
\end{figure}

\begin{figure*}[!t]
    \centering
    \includegraphics[width=1.0\textwidth]{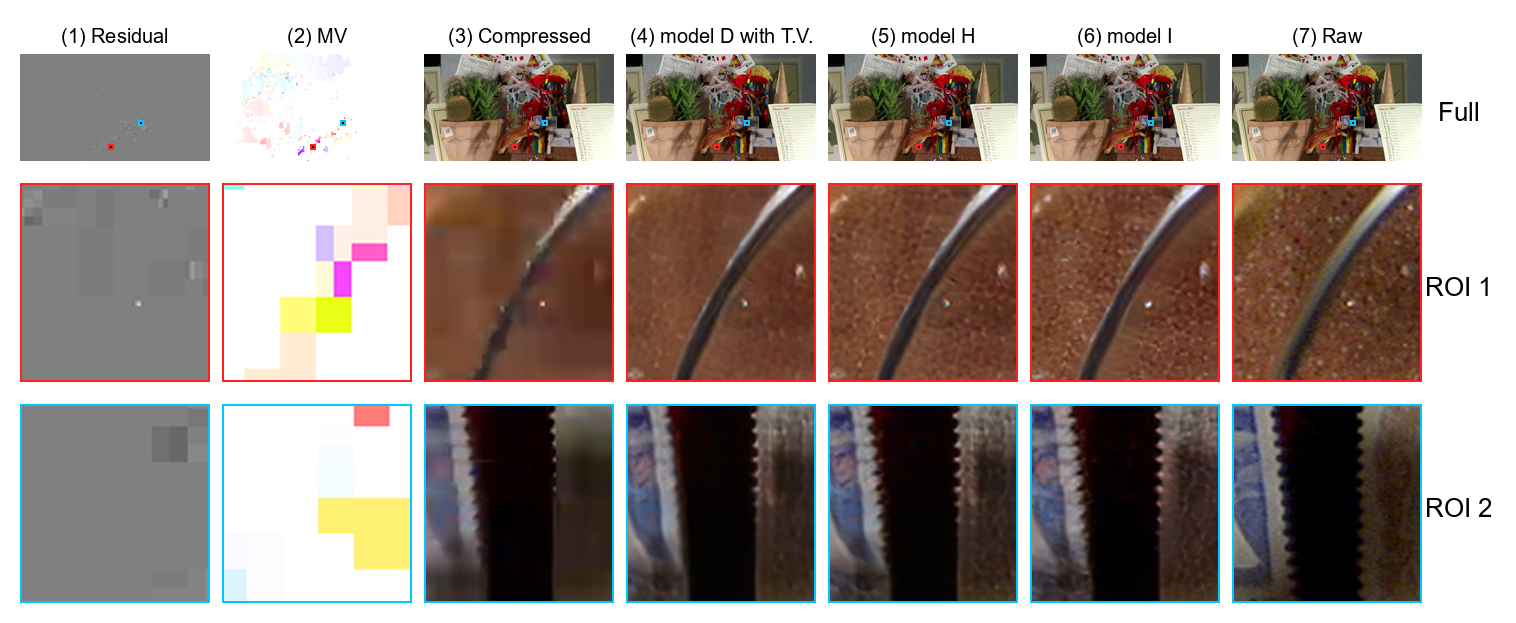}
    \caption{Visualization of the effect of coding priors on perceptual quality enhancement and structure consistency. The first and second columns show the residual and motion-vector priors, respectively. The third column is the compressed frame. The fourth column corresponds to the variant without using coding priors. The fifth column corresponds to the variant with coding priors injected only into the U-Net. The sixth column further fuses coding prior information in the decoder. The last column is the original uncompressed frame.}
    \label{fig:prior_ablation_vis}
\end{figure*}

\textbf{Effectiveness of CDSP:} As shown in Table~\ref{tab:ablation_main}, we evaluate CDSP by comparing models A, B, and \rev{D}. Starting from baseline A, model \rev{D} incorporates the proposed degradation semantic textual prompt and achieves consistent gains. To isolate the effect from lightweight fine tuning, model B is trained with LoRA only. The fact that model \rev{D} still outperforms B demonstrates that the improvements stem from CDSP’s QP-conditioned degradation semantics guidance, rather than LoRA alone. \rev{Additionally, we compare model D with model C, which uses a learnable numerical QP embedding. Model D shows an overall advantage in perceptual quality, further supporting the effectiveness of the proposed textual conditioning.}

Fig.~\ref{fig:CDSP_prompt_vis} visualizes the role of the degradation semantic textual prompt. Compared with the blank prompt setting, our prompt induces distinct cross attention patterns, showing that injected semantics effectively modulate the pretrained diffusion model and enable better adaptation of restoration behavior across compression levels.

\rev{\textbf{Comparison of Text Prompt:} We compare different text prompt settings on Class C and D sequences. Model J is a version of model D trained with the alternative text prompt in Table~\ref{tab:alternative_prompts}. Table~\ref{tab:prompt_variants} compares model B with a blank prompt, model D with the original text prompt in Table~\ref{tab:prompt}, and model J. Both model D and model J improve all five metrics over model B at each QP. Model J yields better $\Delta$MUSIQ and $\Delta$NIQE, while model D achieves better $\Delta$CLIPIQA at all three QPs. Their relative advantages in $\Delta$LPIPS and $\Delta$DISTS depend on the QP. These comparisons show that different text prompts affect the enhancement results differently across metrics. Nevertheless, both model D and model J outperform the blank-prompt baseline, indicating that the benefit of textual conditioning is maintained across the two tested prompt settings.}

\begin{table}[!t]
    \caption{Quantitative comparison of text prompt settings on Class C and D sequences. Best and second-best values for each QP and metric are marked in bold and underlined, respectively.}
    \label{tab:prompt_variants}
    \centering
    \footnotesize
    \renewcommand{\arraystretch}{1.15}
    \setlength{\tabcolsep}{2pt}
    \resizebox{\columnwidth}{!}{\begin{tabular}{c c l c c c c c}
    \toprule
    QP & Model & Setting & $\Delta$LPIPS$\downarrow$ & $\Delta$DISTS$\downarrow$ & $\Delta$MUSIQ$\uparrow$ & $\Delta$NIQE$\downarrow$ & $\Delta$CLIPIQA$\uparrow$ \\
    \midrule
    32 & B & Blank prompt & 0.007 & -0.023 & 4.572 & -0.679 & 0.173 \\
    32 & D & Original prompt & \best{0.004} & \best{-0.025} & \second{4.887} & \second{-0.709} & \best{0.179} \\
    32 & J & Alternative prompt & \second{0.006} & \second{-0.024} & \best{5.263} & \best{-0.719} & \second{0.177} \\
    \midrule
    37 & B & Blank prompt & -0.038 & -0.050 & 8.745 & -1.159 & 0.230 \\
    37 & D & Original prompt & \best{-0.042} & \best{-0.053} & \second{9.760} & \second{-1.247} & \best{0.244} \\
    37 & J & Alternative prompt & \second{-0.040} & \second{-0.051} & \best{9.935} & \best{-1.272} & \second{0.238} \\
    \midrule
    42 & B & Blank prompt & -0.082 & -0.075 & 16.807 & -1.731 & 0.292 \\
    42 & D & Original prompt & \second{-0.085} & \second{-0.077} & \second{17.727} & \second{-1.828} & \best{0.307} \\
    42 & J & Alternative prompt & \best{-0.087} & \best{-0.079} & \best{17.953} & \best{-1.932} & \second{0.301} \\
    \bottomrule
    \end{tabular}}
\end{table}

\begin{table}[!t]
    \caption{Alternative text prompt at each QP.}
    \label{tab:alternative_prompts}
    \centering
    \footnotesize
    \renewcommand{\arraystretch}{1.15}
    \setlength{\tabcolsep}{4pt}
    \begin{tabular}{c >{\raggedright\arraybackslash}p{0.79\columnwidth}}
    \toprule
    QP & Alternative text prompt \\
    \midrule
    32 & a medium-quality video frame degraded by moderate compression, exhibiting mild block artifacts, subtle ringing, barely perceptible blur, and minimal noise \\
    37 & a low-quality video frame degraded by heavy compression, exhibiting noticeable block artifacts, visible ringing, slight blur, and minor noise-like distortions \\
    42 & a very low-quality video frame degraded by severe compression, exhibiting severe block artifacts, pronounced ringing, clear blur, and visible noise-like distortions \\
    \bottomrule
    \end{tabular}
\end{table}

\rev{\textbf{Generalization Across QP Levels:} We evaluate QP generalization on Class C and D sequences at QP 34, 39, 44, 47, 49, and 51. Table~\ref{tab:unseen_qp} compares model B with a blank prompt and model D using the anchor-based embedding interpolation strategy in Eq.~\ref{eq:qp_anchor_interpolation}. Model D achieves lower $\Delta$LPIPS and higher $\Delta$MUSIQ and $\Delta$CLIPIQA at all evaluated QPs. It also achieves lower $\Delta$DISTS at most QPs, with a slight exception at QP 51. These results indicate that CDSP retains its perceptual enhancement benefit across a broader QP range.}

\begin{table}[!t]
    \caption{Generalization across QP levels on Class C and D sequences. Best values for each QP and metric are marked in bold.}
    \label{tab:unseen_qp}
    \centering
    \footnotesize
    \renewcommand{\arraystretch}{1.15}
    \setlength{\tabcolsep}{1.5pt}
    \resizebox{\columnwidth}{!}{
    \begin{tabular}{c c l c c c c}
    \toprule
    QP & Model & Conditioning & $\Delta$LPIPS$\downarrow$ & $\Delta$DISTS$\downarrow$ & $\Delta$MUSIQ$\uparrow$ & $\Delta$CLIPIQA$\uparrow$ \\
    \midrule
    34 & B & Blank prompt & -0.009 & -0.033 & 5.537 & 0.188 \\
    34 & D & Anchor interpolation & \best{-0.014} & \best{-0.036} & \best{6.595} & \best{0.205} \\
    \midrule
    39 & B & Blank prompt & -0.055 & -0.060 & 10.896 & 0.250 \\
    39 & D & Anchor interpolation & \best{-0.058} & \best{-0.062} & \best{14.342} & \best{0.297} \\
    \midrule
    44 & B & Blank prompt & -0.091 & -0.083 & 18.976 & 0.290 \\
    44 & D & Anchor interpolation & \best{-0.096} & \best{-0.087} & \best{22.558} & \best{0.330} \\
    \midrule
    47 & B & Blank prompt & -0.116 & -0.098 & 26.054 & 0.283 \\
    47 & D & Anchor interpolation & \best{-0.121} & \best{-0.099} & \best{28.654} & \best{0.312} \\
    \midrule
    49 & B & Blank prompt & -0.110 & -0.096 & 25.669 & 0.228 \\
    49 & D & Anchor interpolation & \best{-0.114} & \best{-0.097} & \best{26.101} & \best{0.238} \\
    \midrule
    51 & B & Blank prompt & -0.096 & \best{-0.085} & 20.397 & 0.166 \\
    51 & D & Anchor interpolation & \best{-0.107} & -0.084 & \best{21.442} & \best{0.169} \\
    \bottomrule
    \end{tabular}}
\end{table}

\textbf{Effectiveness of CPDC:} As shown in Table~\ref{tab:ablation_main}, we evaluate CPDC using models \rev{D}, \rev{E}, \rev{F}, and \rev{G}. Model \rev{E}, which uses residuals alone, improves performance under stronger compression but becomes less stable at higher bitrates, likely because mildly compressed frames already contain sufficient structural information, making residual guidance partially redundant and harder to exploit under the same training budget. Further introducing motion vectors in model \rev{G} yields consistent gains across all bitrate settings, indicating that effective coding priors should jointly capture spatial error and motion information. Further comparing models \rev{F} and \rev{G}, both use residuals and motion vectors as coding prior conditions, but only \rev{G} applies coding prior modulation in CVCM, leading to consistent improvements in $\Delta$LPIPS and $\Delta$DISTS across QP levels. This indicates that coding prior modulation in CVCM helps correct artifact-biased responses in compressed video features and generate more structurally reliable video conditions, which motivates the final CPDC design.

Fig.~\ref{fig:prior_ablation_vis} further illustrates the effect of coding priors. The residual and motion vector priors in the first two columns highlight local distortion and motion regions, providing cues for restoring compressed details. \rev{Compared with the fourth column, model H in the fifth column better restores fine textures in ROI 1 and more clearly recovers the serrated contour of the stamp edge in ROI 2, producing results closer to the raw frame.} This indicates that U-Net-side coding prior guidance imposes structural constraints during diffusion denoising, improving alignment between generated details and codec-derived residual and motion cues.

\textbf{Effectiveness of CPWF:} As shown in Table~\ref{tab:ablation_main}, we evaluate CPWF using models \rev{G}, \rev{H}, and \rev{I}. Model \rev{H} enhances the decoder with temporal convolution and VAE encoder feature modulation, while model \rev{I} further introduces CPWF. Compared with \rev{H}, model \rev{I} consistently improves $\Delta$LPIPS and $\Delta$DISTS across all QP levels. This shows that decoder-side fusion of coding priors provides complementary information beyond temporal enhancement and VAE modulation, improving perceptual quality. Since the decoder reconstructs pixel-level outputs from latent features, injecting coding priors at this stage further constrains local details and helps preserve structural consistency after denoising.

Fig.~\ref{fig:prior_ablation_vis} further supports this observation. \rev{Compared with model H in the fifth column, model I in the sixth column benefits from decoder-side coding-prior fusion, further enhancing texture details.} These results indicate that CPWF helps the decoder better exploit coding priors to refine local structures after diffusion denoising, reducing remaining structure-inconsistent artifacts beyond U-Net-side guidance.

\section{Conclusion}
\label{sec:conclusion}
In this paper, DiffCVE, a diffusion-based compressed video enhancement method, has been presented. The proposed method incorporates CDSP to adapt the diffusion model to different compression severities through textual guidance for degradation semantics, CPDC to inject coding priors into the denoising process, and CPWF to further exploit coding priors in the decoder. Extensive experiments demonstrate that DiffCVE achieves strong perceptual quality enhancement, especially under severe compression, while also improving temporal consistency. These results verify the effectiveness of jointly leveraging diffusion priors and coding priors for perceptual quality enhancement of compressed video. Despite these promising results, the proposed method still requires more computation and inference time than lightweight non-diffusion enhancement methods, although the single-step variant substantially improves its efficiency. In future work, we will explore more efficient architectures and inference strategies to further reduce the computational cost while achieving further improvements in enhancement performance.

\bibliographystyle{IEEEtran}
\bibliography{references}

\end{document}